\documentclass[journal]{IEEEtran}
\usepackage{soul}

\usepackage{comment}

\usepackage{cite}

\usepackage{enumitem}

\ifCLASSINFOpdf
  \usepackage[pdftex]{graphicx}
  \graphicspath{{images/}}
\else
\fi
\usepackage[dvipsnames]{xcolor}

\usepackage{amssymb}	
\usepackage{mathtools,empheq}
\usepackage{color}

\newcommand{\vect}[1]{\boldsymbol{\mathbf{#1}}}
\newcommand{\M}{\mathcal{M}}
\newcommand{\F}{\mathcal{F}}
\newcommand{\R}{\mathbb{R}}			
\newcommand{\T}{^{\mathrm{T}}}		
\DeclareMathOperator{\atan}{atan}
\newcommand{\norm}[1]{\left\lVert#1\right\rVert}
\newcommand{\abs}[1]{\left|#1\right|}
\newenvironment{nalign}{
    \begin{equation}
    \begin{aligned}
}{
    \end{aligned}
    \end{equation}
    \ignorespacesafterend
}

\usepackage{algorithmic}
\usepackage{algorithm}

\usepackage{siunitx}

\usepackage{subcaption}
\usepackage{float}
\usepackage{placeins}

\usepackage{multirow}
\usepackage{booktabs} 
  \newcommand{\ra}[1]{\renewcommand{\arraystretch}{#1}}
  \newcolumntype{G}{>{\centering}p{3pt}}
  \newcommand{\bigcell}[2]{\begin{tabular}{@{}#1@{}}#2\end{tabular}}

\usepackage[nameinlink,capitalise]{cleveref}

\begin{document}
%
\title{ Autonomous Cooperative Flight \\ Control for Airship Swarms}

%
%
%

\author{Pedro G. Artaxo,
		Auguste~Bourgois,
        Hugo~Sardinha,
        Henrique Vieira,
        Ely~Carneiro~de~Paiva,
        Andr\'{e}~R.~Fioravanti and Patricia~A.~Vargas,~\IEEEmembership{Member,~IEEE}
\thanks{Auguste Bourgois, Hugo Sardinha and Patricia A. Vargas are with the Robotics Laboratory, Edinburgh Centre for Robotics, School of Mathematical and Computer Sciences, Heriot-Watt University, Edinburgh, UK EH144AS e-mail: p.a.vargas@hw.ac.uk.}


\thanks{Pedro Artaxo, Ely Carneiro de Paiva, Andr\'{e} Fioravanti and Henrique Vieira are with the University of Campinas (UNICAMP), Campinas, Brazil, e-mail: elypaiva@fem.unicamp.br}

\thanks{Manuscript received March day, 2018; revised Month day, 2018.}}

\maketitle

\begin{abstract}
This work investigates two approaches for the design of autonomous cooperative flight controllers for airship swarms. The first controller is based on formation flight and the second one is based on swarm intelligence strategies.  In both cases, the team of airships needs to perform two different tasks: waypoint path following and ground moving target tracking. The UAV platform considered in this work is the NOAMAY airship developed in Brazil. We use a simulated environment to test the proposed approaches. Results show the inherent flexibility of the swarm intelligence approach on both tasks.
\end{abstract}


%
\IEEEpeerreviewmaketitle

\section{Introduction}
\label{sec:introduction}

\IEEEPARstart{T}{he} development of unmanned aerial vehicles (UAVs) for use as single units or in collaborative robotic swarms constitutes an important and emergent area of scientific and technological research. Compared to traditional manned aircrafts, they are less expensive, more flexible to operate and do not require an on-board human pilot \cite{Kumar2, Mahony2012}. This helps to explain the rapid growth of applications of UAVs in agricultural scenarios like crop and live stock monitoring, overall environmental monitoring, forest fire detection,  inspection, surveillance, convoy protection and search and rescue \cite{tsourdos2010cooperative, oh2013towards}.

Within the environmental context, UAVs have appeared as a promising alternative for mapping, monitoring, supporting and surveillance applications in rangelands, forests and preservation areas \cite{Salami2014, Bueno2002, Carvalho2014, Simon2017}. For instance, UAV-based applications yield a much better resolution (from hundreds of meters to centimeters) for remote sensing when compared to satellite-based or full-scale manned aircraft. Therefore, UAVs may be used to rapid response needs and can be equipped with a greater variety of sensors, beyond the usual multispectral imaging, as they can fly closer to the ground \cite{Salami2014}. Rainforest areas are also becoming candidates for extended applications of UAVs, such as estimating canopy cover and gap sizes, monitoring of biodiversity, assessing ephemeral erosion and human interference, analysing gaseous components plus estimating biomass that can be transformed into carbon credits \cite{Simon2017}.

UAV platforms are usually classified into three groups: rotor type (helicopters, quadcopters, etc), wing type (planes) or ``lighter-than-air'' (airships and balloons), which in turn will define the basic features on endurance, range, altitude, and aerodynamics profile of the aircraft \cite{Kumar2, Liew}. In the case of airships, the aerostatic lift comes from a lighter-than-air gas (Helium), whereas movement is typically ensured by a pair of controllable propellers (or even two pairs, like in our case) plus the aerodynamic surfaces on the tail \cite{Bueno2002}, \cite{moutinho2016airship}.

Overall, airships are best suited for applications in environmental monitoring and surveillance over other UAVs for their unique features:
(i) low interference on the surrounded environment including low noise generation with the use of electrical motors; (ii) hovering capability, with vertical take-off and landing without the need of runway and, finally (iii) greater flight endurance as they derive the largest part of their lift from aerostatics \cite{Bueno2002, Carvalho2014, Liew}. A medium size airship (about 20-40 m), for example, can stay on the air for hours and a large airship ($>$40 m) for days, while its available payload will depend on the volume of the envelope \cite{bestaoui_book}. However, due to their lateral under-actuation, slower dynamics, non-linearities in the model (including actuators saturation) and greater sensibility to wind disturbances, the development of controllers, guidance and navigation systems for airships is a relatively complex task \cite{moutinho2016airship, bestaoui_book, Paiva2006, Liesk}.

For some kinds of applications, the use of a set of multiple UAVs flying autonomously and cooperatively, as in a swarm, may be more effective than the use of a single UAV \cite{Brambilla, Dasgupta2008, Dong2015}. In this case, each individual UAV can be entrusted with different functions in a given mission (task allocation), or they can divide the same workload with enhanced flexibility (adaptability, scalability, maintainability) and robustness (reliability, sustainability and fault-tolerance) \cite{Kumar2}.

In this context, we propose and investigate two approaches for the design of autonomous cooperative flight controllers of a team of airships.\footnote{This work on airships swarm is a result of the SAS-ROGE project, a collaborative research between University of Campinas (BR) and Heriot-Watt University (UK)} The first one is based on formation flight and the second one is based on swarm intelligence strategies.  In both cases, the team of airships should execute two different tasks: (i) waypoint path following; and (ii) moving target tracking. The UAV platform considered in this work is the NOAMAY airship \cite{Ruedas2017, Vieira2017}, a prototype that is a result of two Brazilian projects: the DRONI and the InSAC Projects. The 10m long aircraft, shown in Fig. 1b, is an improved design version of the pioneer airship of Project AURORA - ``Autonomous Unmanned Remote Monitoring Robotic Airship'' \cite{Bueno2002, Paiva2006}, shown in Fig. 1a. This new configuration has four electrical tilting propellers with improved features of controllability and maneuverability.\footnote{The NOAMAY airship will be used in a pilot experiment for surveillance/monitoring in the Amazon rainforest, more specifically at the Mamiraua Reserve in partnership with the Mamiraua Sustainable Development Institute-IDSM \cite{Carvalho2014}.}

\begin{figure}[!thb]
	\centering
	\includegraphics[width=0.7\linewidth]{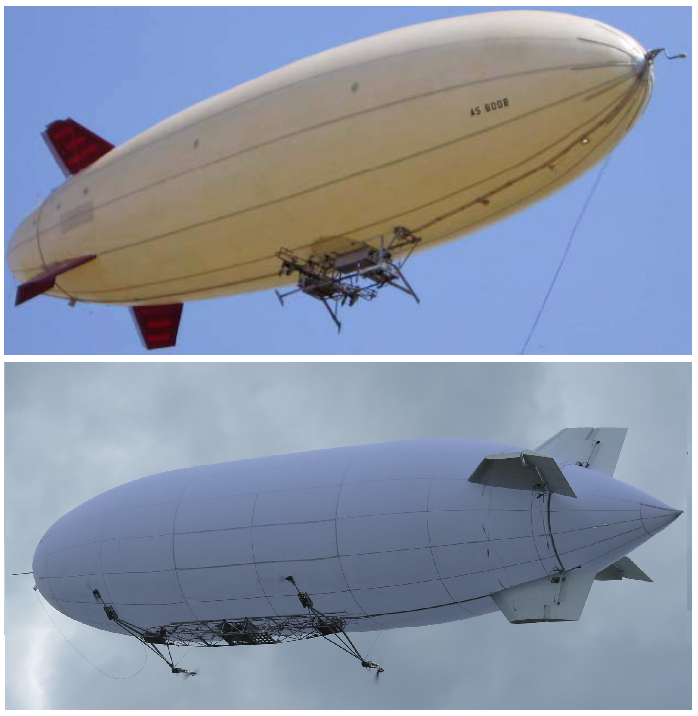}
	\caption{Top: pioneer AURORA airship used in different research investigations. Bottom: NOAMAY airship platform (2018), whose flight simulator is used in this work.}
	\label{fig:platforms}
\end{figure}

In the past few years, the problem of cooperative flight control of UAVs has already been extensively investigated in the scientific literature for fixed wing and rotor type UAVs \cite{Kumar2, oh2013towards, Kumar1}, including the problem of detection and following of moving targets on the ground \cite{Encarna, Dixon}. Nonetheless, for the lighter-than-air platforms there are only few works in the literature, mainly focused on the use of multiple airships or balloons for {\it{indoor}} applications in a friendly environment, where there are almost no wind disturbances  \cite{Olsen1999, King2004}. And to the best of the authors knowledge, the only work in the literature focusing on cooperative flight of {\it{outdoor}} airships is \cite{bicho2006airship}, where underactuation and wind perturbations impose special challenges. However, this   particular study presents some important drawbacks. Firstly, it is restricted to the waypoint path-following case, while more complex mission cases like hovering flight and moving target tracking were not considered. Moreover, the simulation signals of the control actuators, aerodynamic variables, attitude angles and position/velocities of the airships were not clearly presented, making it more difficult to analyse and assess the behaviour, performance and robustness of the proposed approach.

Active research has also been developed in the last years related to surveillance and monitoring with UAVs, including efficient algorithms for data acquisition/analysis \cite{tsourdos2010cooperative, oh2013towards},   information and sensor fusion \cite{Salami2014, Dixon}, low weight sensor/embedded systems \cite{Liew}, dynamic mission planning \cite{Kumar1, Encarna}, automatic UAV coordination \cite{Dong2015, King2004, Olsen1999} and swarm algorithms \cite{Kumar2}.  In this paper, we focus in automatic UAV coordination and swarm algorithms, designing and testing autonomous coordination controllers (via formation flight or intelligent swarm) for a team of airships performing two tasks: waypoint path-following (including eventual hovering over a point of observation) and tracking of a moving target on the ground.

The remaining of this paper is organised as follows. Section II presents the problem formulation with the basic assumptions, objectives and methodologies. Section III describes the basic kinematic/dynamic models of the airship platform used in this work. Section IV highlights the fundamentals of intelligent swarm design. Section V presents the fundamentals of the formation flight design. Section VI shows results from both approaches on the two mission tasks. Finally, design approaches and results are discussed in Section VII and in Section VIII we draw conclusions and propose future work.

\section{Problem Formulation and Methodology}
\label{sec:probformulation}

The first task to be performed by the team of airships is the \textit{waypoint path following} that can be used for different purposes (Fig 2(a)). Usually waypoint path following is used for surveillance/inspection of selective points and area coverage (i.e., a systematic and regular search over a given investigated terrain).  In order to increase the efficiency of the mission, a number of factors like: the payload, the set of sensors used, and the duration of the mission, can impose important limitations that motivate the use of multiple UAVs \cite{Kumar2}. In the case of waypoint path following, the team of airships receives a set of target coordinates to visit, flying at a given altitude and speed. Upon arriving at a point, the airship may proceed to the next point, or it may stay in place (hovering), over a chosen radius circle from the targeted point, thus enabling the airship to face the wind (a technical constraint necessary to minimize drag and increase maneuverability) \cite{azinheira2006airship}.


\begin{figure}[!htb]
    \centering
    \includegraphics[width=0.99\linewidth]{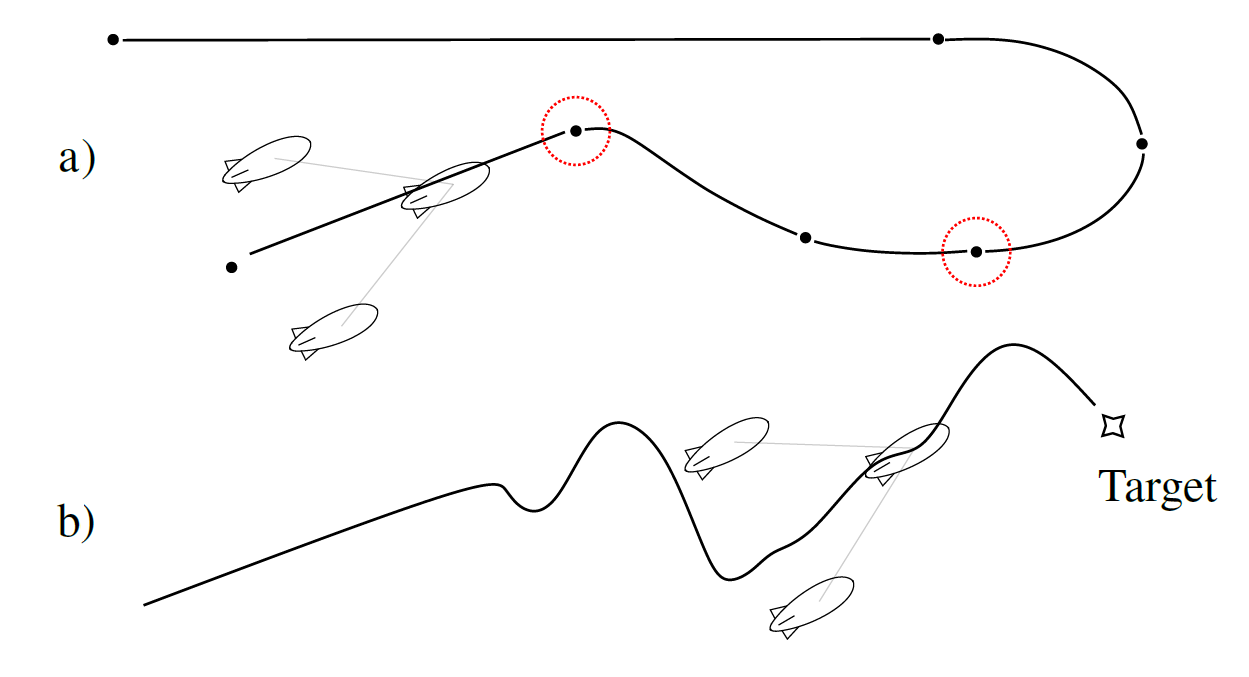}
    \caption{Tasks: (a) \textit{waypoint path following} with eventual hovering flight - red circles, and (b) \textit{tracking of moving target}.}
    \label{fig:mission_tasks}
\end{figure}

The second task is \textit{tracking of moving targets} (Fig 2(b)) like livestock, wild animals, people or other vehicles on the ground, which is very complex due to the unforeseen maneuvers of the target, as well as the kinematic constraints of the aircraft \cite{Encarna}, \cite{Dixon}. In this case, a cooperative flight of a team of UAVs may benefit from a wider search field and from a better estimation of the target position due to the sensor readings fusion that comes from the other UAVs \cite{oh2013towards}.



The problem of using a set of UAVs to track a trajectory or a moving target becomes increasingly complex if the team has also to obey a given flight formation or interaction protocol. This turns the problem into an even more challenging task if we consider the difficulties associated to the control of outdoor airships \cite{moutinho2016airship, Vieira2017, azinheira2006airship}. We propose here two different approaches to solve this problem: formation flight \cite{Dong2015, bicho2006airship} and swarm intelligence \cite{Kumar2}, whose basic concepts and principles are presented below.

\subsection{Formation Flight}
\label{subsec:formation}

The first approach used here is the classical cooperative formation flight \cite{Dong2015, Olsen1999, bicho2006airship}. This strategy is usually based on the classical hierarchy of communication and coordination \cite{tsourdos2010cooperative} of a cooperative mission planning (Fig 3). The top layer defines the mission task allocation, the middle layer is the guidance block that defines the routes/trajectories/references to be followed by each UAV, and the lower layer corresponds to the local controllers that ensure that the UAVs will execute the commanded trajectories/velocities.

\begin{figure}[!thb]
	\centering
	\includegraphics[width=0.98\linewidth]{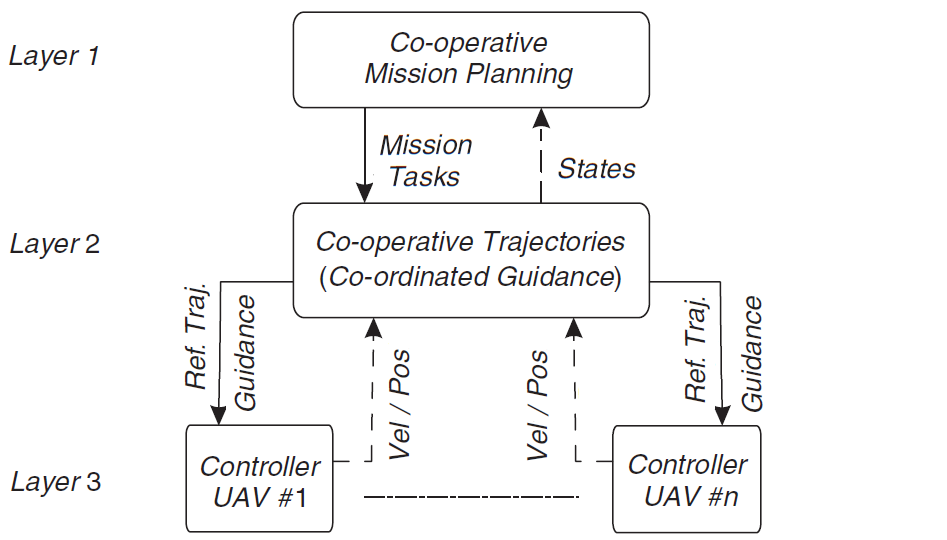}
	\caption{Typical hierarchy of mission planning in coordinated flight adapted from \protect\cite{tsourdos2010cooperative}.}
	\label{fig:architecture}
\end{figure}

In this work, we focus our investigation in layers 2 (path planning and guidance) and 3 (low level control), as we suppose that the flyable paths or waypoints are given and no optimisation on the route/path is executed. However, regarding the path planning, which is a higher level component of the middle layer, we briefly consider it here, for example to generate a better way to approach and stay at a given distance from the searched object in the target tracking task. 

The coordinated formation approach used here is the leader/follower technique, very common on UAVs coordinated flight \cite{Kumar2, Dong2015, bicho2006airship}. In this mode, the follower airship should maintain a desired distance and orientation angle from the leader, which allows for a flying formation to be achieved, be it in line, column or diagonal (“V" formation). It is implemented here using a State Feedback Kinematic Control (SFKC) approach for layer 2 (guidance), commonly used in mobile robotics for terrestrial vehicles \cite{siegwart2011introduction}. And for the low level (layer 3) we consider an adaptation of the Nonlinear Sliding Mode Control (SMC) of \cite{Vieira2017}, whose purpose is to track the commanded position/orientation/velocity of the autonomous vehicles. 


\subsection{Swarm Intelligence}
\label{subsec:swarm}

In the 1980’s, Beni, Hackwood, and Wang introduced the concept of \textit{swarm intelligence} \cite{bonabeau1999swarm} that can be defined as the collective intelligence \cite{tan2013research} emerging from a group of simple agents. Many observed animal species are known to display \textit{swarm intelligence} behaviour. These behaviours are generally classified as follows \cite{bonabeau1999swarm, sardinha2016swarm, kingma2017walter}:
\begin{itemize}
\item Flocking: a flocking behaviour occurs when animals gather and move together as what could seem to be a single organism (e.g, a flock of starlings or a school of mackerels).
\item Foraging: a foraging behaviour occurs when animals collaborate to exploit their environment (e.g., ants of a colony share the work of harvesting food all around the nest).
\item Collaborative Manipulation: some animals sometimes gather to carry heavy loads (e.g, ants are also known to have such behaviour).
\item Aggregation: some animals sometimes regroup and stay together (e.g., emperor penguins do so to protect themselves from the cold).
\end{itemize}

The study of swarm intelligence gave rise to many biologically inspired optimization algorithms that usually can provide better performance than classical approaches \cite{bonabeau1999swarm}. The presence of collective intelligence in nature emerges from the lack of a global communication system, for instance, in flocks of birds, schools of fish or ant colonies. Computer scientists have been exploring approaches to mimic this collective intelligence by creating strategies that would resemble the behaviour of these kinds of insects and animals. A swarm robotics approach makes use of one or more of these strategies that mimic collective intelligence in order to coordinate a group or swarm of robots. The simple robots that form the swarm are able to communicate with one another, either explicitly, implicitly or through passive action recognition \cite{DAS201646}, and usually have a stochastic behaviour that can generate some sort of collective intelligence \cite{trianni2008evolutionary, turgut2008self, Brambilla}.

The main characteristics of swarm robotics are:

\begin{itemize}
\item Robustness: swarm robots are able to continue working and accomplish their mission despite suffering the failure of some of their individual members. This capability is a result of the decentralised distributed architecture that characterises swarms.
\item Flexibility: these kinds of systems can adapt easily to changes in the requirements as well as to perturbations in the environment.
\item Scalability: given that swarm robotic systems are based on decentralised distributed systems, it is possible to increase the number of agents without sacrificing performance. The computational complexity does not increase with the number of agents since each of them relies on its own independent controller.
\item Emergence: the most important property of these kinds of systems is the emergence of a collective intelligent behaviour from simple interactions between the local agents (swarm members) and the environment. Usually each agent follows a set of simple rules, which results into an intelligent behaviour as a group.
\end{itemize}

Swarm robotics approaches have been applied to a range of applications for optimisation and resource-allocation tasks, including the traveling-salesman problem, job-shop scheduling, vehicle routing, self-repairing formation control for mobile agents, load balancing, searching, autonomic computing in overlay networks and flight coordination \cite{Brambilla}. 

In the course of this work two major swarm intelligence strategies are used to establish coordination of a swarm of airships: the Boids Algorithm \cite{reynolds1987flocks} and the Robotic Particle Swarm Optimization (RPSO) \cite{couceiro2011novel}. Both will be detailed in Section V.

\subsection{Experimental Assumptions}

 In this work we assume that each airship is provided with the necessary equipment/sensors and network communications for a precise navigation as well as for the recognition and tracking of the moving target. With regards to the mission architecture type, the formation flight is a semi-decentralized architecture (e.g., only the ``leader'' UAV receives external commands), while the intelligent swarm is a decentralized architecture (e.g., the UAVs are distributed autonomously in relation to their closest neighbours) \cite{tsourdos2010cooperative, Dasgupta2008}. 

The development, test and validation of the approaches will be restricted to the airship computational simulation environment, a 6-DOF dynamic simulator (Simulink/Matlab), based on a high fidelity dynamic model including wind/turbulence and aerodynamic coefficients from wind tunnel experiments \cite{moutinho2016airship, Paiva2006, gomes1998airship}. For the formation flight controller we have used the full nonlinear dynamic model, while for the intelligent swarm controller we used a simplified version of the simulator considering a pseudo-kinematic model \cite{Artaxo2018}. Moreover, for simulation purposes, and considering the heavy computational effort of multiple aircraft models, we consider a configuration of 3 airships in the waypoint tracking application and 4 airships in the target tracking case. Finally, for the intelligent swarm approach, we suppose the prior existence of a collision-avoidance that is natural from this technique and for the formation flight we consider a collision-free motion environment \cite{Kumar2}, where each airship flies at a given unique altitude.

\section{Airship Model}
\label{sec:model}

This section presents the full 6-DOF airship model used in the simulation environment, the simplified versions of this dynamic model for control design purposes, and the kinematic airship model.

\vspace{-0.2cm}

\subsection{Full Airship Dynamic Model and Simulator}
\label{subsec:act}
The Noamay airship model/simulator (in Simulink/Matlab) is an evolution of the previous AURORA airship  model/simulator \cite{moutinho2016airship, Paiva2006}, incorporating the new propulsion system of 4-electrical tilting thrusters \cite{Vieira2017}, and replicating multiple airship models in a same simulation environment under independent wind/turbulence conditions.

\begin{figure}[!htb]
    \centering
    \includegraphics[width=0.98\linewidth]{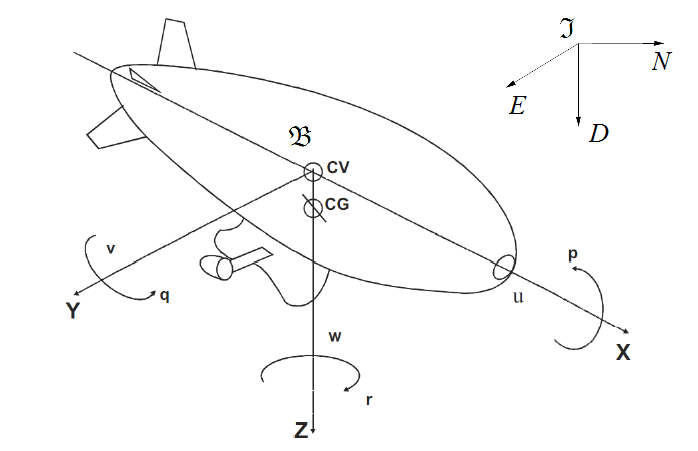}
    \caption{Airship's axis frames and main variables. }
    \label{fig:axis}
\end{figure}		

The core of this airship model/simulator is the nonlinear and non-deterministic equation (with a stochastic wind/gust model) obtained from the Newton's first law, that is:
\begin{equation}
	\M
	\begin{bmatrix}
		\dot{\vect{v}}     \\
		\dot{\vect{\omega}}
	\end{bmatrix} = \F_{d} + \F_{a} + \F_{p} + \F_{w} + \F_{g} \label{eq: generalized dynamics}
\end{equation}
where $ \M \in \R^{6\times6}$ is the mass and generalised inertia matrix (including virtual mass and inertia effects) and $ \vect{v} \in \R^{3} $ and $ \vect{\omega} \in \R^{3} $ are the vectors of inertial and angular velocities in the local reference frame. The vectors of forces/moments on the right side of the equation are: $ \F_{d} $ - centrifugal/Coriolis forces; $ \F_{a} $ - aerodynamic forces; $ \F_{p} $ - propulsion forces; $ \F_{w} $ - wind forces (due to wind induced dynamics) and, finally, $ \F_{g} $ - gravitational/buoyancy forces \cite{moutinho2016airship, Paiva2006, Vieira2017}. 

\begin{figure}[!htb]
	\centering
	\includegraphics[width=0.98\linewidth]{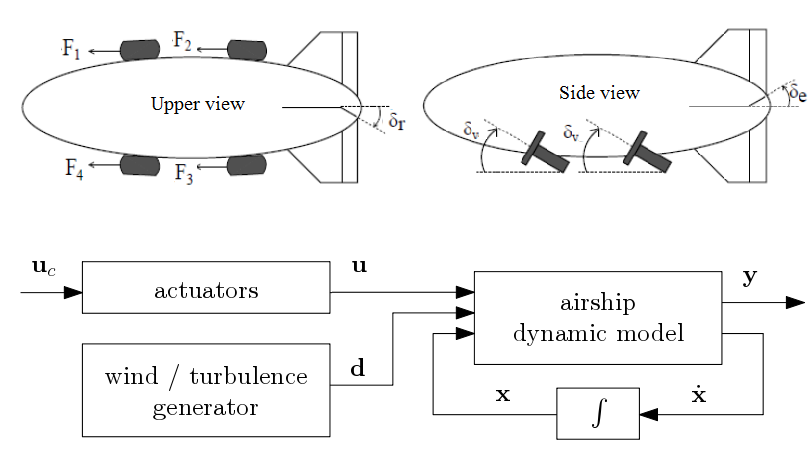}
	\caption{Airship actuators (above) and open-loop dynamic model blocks (below).}
	\label{fig:open_loop}
\end{figure}

The fundamental aspect of the dynamic simulator is the numerical solution of this mathematical model, while incorporating operational aspects like saturation/dynamics of actuators \cite{moutinho2016airship}, aerodynamic model with parameters obtained from wind tunnel experiments \cite{gomes1998airship}, wind speed estimator and others \cite{Paiva2006}. Thus, to derive the airship 6D motion, the simulator solves numerically the differential equation formalized by:

\begin{equation}
    \vect{\dot{\bar{x}}=\vect{f(\bar{x}, u, d)}} 
    \label{eq:eq_simulator}
\end{equation}
where:
\begin{itemize}
\item The state $\vect{\bar{x}} = [\vect{v}\T \ \vect{\omega}\T \ \vect{\Phi}\T \ \vect{p}\T]\T $ includes the linear $\vect{v}=\left[u \ v \ w\right]^T$ and angular $\vect{\omega}=\left[p \ q \ r\right]^T$ inertial velocities of the airship expressed in body-fixed $\mathfrak{B}$ frame, as well as the Euler angles ($ \vect{\Phi} = [\varphi \ \theta \ \psi]\T$) and the Cartesian positions of the CV in the inertial frame $\mathfrak{I}$ ($\vect{p} = [P_N \ P_E \ P_D]\T $) (\cref{fig:axis}).

\item The input vector, given by
$	\vect{u} = [\vect{\delta}_e \ \vect{\delta}_r \ \vect{\delta}_1 \ \vect{\delta}_2 \ \vect{\delta}_3 \ \vect{\delta}_4 \ \vect{\delta}_{v1} \ \vect{\delta}_{v2} \ \vect{\delta}_{v3} \ \vect{\delta}_{v4}]\T
$, includes the tail aerodynamic surface deflections ($\vect{\delta}_e \ \vect{\delta}_r$), the engine thrust inputs ($\vect{\delta}_i$) and the engines vectoring angles ($\vect{\delta}_{vi}$) (\cref{fig:open_loop}).

\item The disturbance vector $\vect{d} \in \mathbb{R}^6$ includes the wind input (wind velocity)  expressed in the  $\mathfrak{I}$ frame with a constant term ($\vect{V}_w\in \mathbb{R}^3$) and a six component vector modelling the atmospheric turbulence (non-constant wind) in Dryden model \cite{moutinho2016airship, Paiva2006}.
\end{itemize}

\subsection{Simplified Dynamic Models for Control Design}

For control design purposes, however, we use two simplified versions of the mathematical model  (\ref{eq: generalized dynamics}). The first one considers the same nonlinear dynamic model except for the aerodynamic forces, that is disregarded, yielding a deterministic model (no gust). The second one is the linearised version of this model, that also helps to decouple the equations of motion into lateral (\cref{eq:latModel}) and longitudinal (\cref{eq:longModel}) models:
\begin{equation}\label{eq:latModel}
	\begin{bmatrix}
		\dot{\tilde{v}}       \\
		\dot{\tilde{p}}       \\
		\dot{\tilde{r}}       \\
		\dot{\tilde{\varphi}}
	\end{bmatrix}
	= A_{lat}
	\begin{bmatrix}
		\tilde{v}       \\
		\tilde{p}       \\
		\tilde{r}       \\
		\tilde{\varphi}
	\end{bmatrix} + B_{lat}
	\begin{bmatrix}
		\tilde{\vect{\delta}}_r    \\
		\tilde{\vect{\delta}}_{cd}
	\end{bmatrix}
\end{equation} 
\begin{equation}\label{eq:longModel}
\begin{bmatrix}
	\dot{\tilde{u}}       \\
	\dot{\tilde{w}}       \\
	\dot{\tilde{q}}       \\
	\dot{\tilde{\theta}}
\end{bmatrix}
= A_{long}
\begin{bmatrix}
	\tilde{u}      \\
	\tilde{w}      \\
	\tilde{q}      \\
	\tilde{\theta}
\end{bmatrix} + B_{long}
\begin{bmatrix}
	\tilde{\vect{\delta}}_e    \\
	\tilde{\vect{\delta}}_{tt} \\
	\tilde{\vect{\delta}}_{fb} \\
	\tilde{\vect{\delta}}_{vt}
\end{bmatrix}
\end{equation} 
where the superscript ``tilde'' denotes small variations on the variables around their trimmed values and:
\vspace{0.2cm}

\begin{tabular}{@{$\bullet$ }ll}
	$ \vect{\delta}_e $    & Elevator deflection angle;                                                 \\
	$ \vect{\delta}_r $    & Rudder deflection angle;                                                   \\
	$ \vect{\delta}_{tt} $ & $ \vect{\delta}_1 + \vect{\delta}_2 + \vect{\delta}_3 + \vect{\delta}_4 $; \\
	$ \vect{\delta}_{fb} $ & $ \vect{\delta}_1 - \vect{\delta}_2 - \vect{\delta}_3 + \vect{\delta}_4 $; \\
	$ \vect{\delta}_{cd} $ & $ \vect{\delta}_1 - \vect{\delta}_2 + \vect{\delta}_3 - \vect{\delta}_4 $  \\
	$ \vect{\delta}_{tv} $ & Thrust Vectoring = $ \vect{\delta}_{vi} $;                     \\ 
	$ A_{lat},B_{lat}$ & Matrices of lateral model;                     \\  
	$ A_{long},B_{long}$ & Matrices of the long. model;                     \\ 
\end{tabular}
\vspace{0.2cm}
\\
Both simplified models are fundamental for the low level controller design, as will be shown further in Section V.

\subsection{Airship Kinematic model}
\label{sec:kin}	
We introduce here the airship kinematic model that is particularly useful to describe the airship's behaviour at low airspeed velocities. The kinematic equations are based on the transformation matrices relating linear and angular velocities represented in local and inertial frames. To represent the airship orientation (attitude), we adopt here the quaternions approach, instead of the Euler angles \cite{moutinho2016airship}. Let us define the pose vector of the airship as  $\vect{\eta}=\left[\vect{p}^T \ \vect{q}^T\right]^T \in \mathbb{R}^7$ as being composed by Cartesian coordinates $\vect{p}\in \mathbb{R}^3$ in the inertial (NED - \textit{North, East, Down}) frame, and the orientation described by the quaternions $\vect{q} \in \mathbb{R}^4$ \cite{moutinho2016airship}. Therefore, the derivatives of the position (NED) and the quaternions vector can be written, respectively, as:

\begin{nalign}\label{eq:7D-kinematics-1}
    \dot{\vect{p}} & = \vect{S}^{T}\vect{v}    \\
    \dot{\vect{q}} & = \frac{1}{2} \vect{Q} \begin{bmatrix*} 0 \\ \vect{\omega} \end{bmatrix*}
\end{nalign}
where $\vect{S}\in\mathbb{R}^{3\times3}$ is the transformation matrix converting linear velocities from inertial to local frames, whose elements are functions of the quaternions, as well as the unitary matrix $\vect{Q}\in\mathbb{R}^{4\times4}$ that relates the quaternions derivatives to the airship angular velocities \cite{azinheira2006airship}. Therefore, defining  $\vect{x}=[\vect{v} \ \vect{\omega}]^T =\left[u \ v \ w \ p \ q \ r\right]^T$, the derivative of the airship pose vector $\vect{\eta}$ can be written as:
\begin{nalign}
    \dot{\vect{\eta}} &= \begin{bmatrix*} \dot{\vect{p}} \\ \dot{\vect{q}} \end{bmatrix*} = \begin{bmatrix*}
        \vect{S}^{T} & \vect{0}            \\
            \vect{0} & \frac{1}{2}\vect{Q}
    \end{bmatrix*} \vect{C} 
    \begin{bmatrix*} \vect{v} \\ \vect{\omega} \end{bmatrix*} \\
    &= 
    \begin{bmatrix*}
        \vect{S}^{T} & \vect{0}           \\
            \vect{0} & \frac{1}{2}\vect{Q}
    \end{bmatrix*}
    \vect{Cx} = \vect{DCx} = \vect{Tx}	
\end{nalign}	
where 
\begin{nalign}
    \vect{C} &=
    \begin{bmatrix*}
    	  \vect{I}_{3} &   \vect{0}_{3} \\
    	\vect{0}_{1,3} & \vect{0}_{1,3} \\
       	  \vect{0}_{3} &   \vect{I}_{3}
    \end{bmatrix*} &\in\mathbb{R}^{7\times6} \\
    \vect{D} &=
    \begin{bmatrix*}
    	\vect{S}^{T} & \vect{0}           \\
    	\vect{0}     & \frac{1}{2}\vect{Q}
    \end{bmatrix*} &\in\mathbb{R}^{7\times7}\\
    \vect{T} &= \vect{DC} &\in\mathbb{R}^{7\times6}.
\end{nalign}

It is important to recall also that the relation between relative airspeed ($\vect{x}_a$), ground speed ($\vect{x}$) and wind speed ($\vect{x}_w$) represented in the local frame is given by:
\begin{equation}
    \vect{x}_a=\vect{x}-\vect{x}_w
\end{equation}
where:
\begin{nalign}
\vect{x}_a=\left[u_a \ v_a \ w_a \ p_a \ q_a \ r_a\right]^T \\
\vect{x}=\left[u~ \ v~ \ w~ \ p~ \ q~ \ r~\right]^T \\  
\vect{x}_w=\left[u_w \ v_w \ w_w \ 0 \ 0 \ 0\right]^T.
\end{nalign}

\section{Formation Flight}
\label{sec:formationflight}

As stated in Section II, in the formation flight design we focus our investigation in layers 2 (path planning and guidance) and 3 (low level control) of the mission planning architecture presented in Fig 3. The design of the automatic controllers of both layers is detailed next.

\subsection{Second Layer - Kinematic Based Guidance}
For the middle layer (guidance), we use here the ``leader-follower'' coordinated formation flight approach. Although extensively investigated for different kinds of aircrafts \cite{Dong2015}, to our knowledge, there is only a single work in the literature focused on coordinated flight of airships, which is based on the ``leader-follower'' approach \cite{bicho2006airship}.

\begin{figure}[!htb]
	\centering
	\def\svgscale{.5}
\begingroup%
  \makeatletter%
  \providecommand\color[2][]{%
    \errmessage{(Inkscape) Color is used for the text in Inkscape, but the package 'color.sty' is not loaded}%
    \renewcommand\color[2][]{}%
  }%
  \providecommand\transparent[1]{%
    \errmessage{(Inkscape) Transparency is used (non-zero) for the text in Inkscape, but the package 'transparent.sty' is not loaded}%
    \renewcommand\transparent[1]{}%
  }%
  \providecommand\rotatebox[2]{#2}%
  \newcommand*\fsize{\dimexpr\f@size pt\relax}%
  \newcommand*\lineheight[1]{\fontsize{\fsize}{#1\fsize}\selectfont}%
  \ifx\svgwidth\undefined%
    \setlength{\unitlength}{404.333424bp}%
    \ifx\svgscale\undefined%
      \relax%
    \else%
      \setlength{\unitlength}{\unitlength * \real{\svgscale}}%
    \fi%
  \else%
    \setlength{\unitlength}{\svgwidth}%
  \fi%
  \global\let\svgwidth\undefined%
  \global\let\svgscale\undefined%
  \makeatother%
  \begin{picture}(1,0.43621129)%
    \lineheight{1}%
    \setlength\tabcolsep{0pt}%
    \put(0,0){\includegraphics[width=\unitlength,page=1]{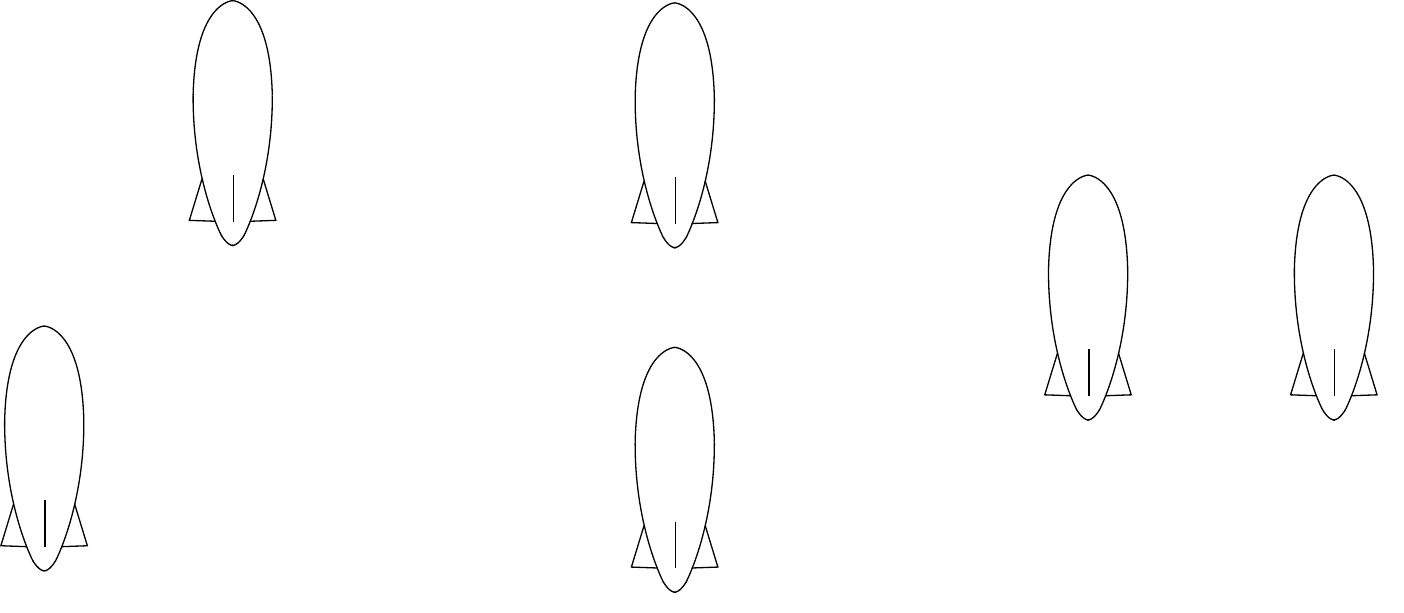}}%
    \put(0.17853194,0.23239228){\color[rgb]{0,0,0}\makebox(0,0)[lt]{\lineheight{1.25}\smash{\begin{tabular}[t]{l}Leader\end{tabular}}}}%
    \put(0.25005743,0.26510727){\color[rgb]{0,0,0}\makebox(0,0)[lt]{\begin{minipage}{0.02384878\unitlength}\raggedright \end{minipage}}}%
    \put(0.51648677,0.24723453){\color[rgb]{0,0,0}\makebox(0,0)[lt]{\lineheight{1.25}\smash{\begin{tabular}[t]{l}Leader\end{tabular}}}}%
    \put(0.89899461,0.1006334){\color[rgb]{0,0,0}\makebox(0,0)[lt]{\lineheight{1.25}\smash{\begin{tabular}[t]{l}Leader\end{tabular}}}}%
    \put(0.04632657,0.00389295){\color[rgb]{0,0,0}\makebox(0,0)[lt]{\lineheight{1.25}\smash{\begin{tabular}[t]{l}Follower\end{tabular}}}}%
    \put(0.51478096,0.00044924){\color[rgb]{0,0,0}\makebox(0,0)[lt]{\lineheight{1.25}\smash{\begin{tabular}[t]{l}Follower\end{tabular}}}}%
    \put(0.69430924,0.1006304){\color[rgb]{0,0,0}\makebox(0,0)[lt]{\lineheight{1.25}\smash{\begin{tabular}[t]{l}Follower\end{tabular}}}}%
    \put(0,0){\includegraphics[width=\unitlength,page=2]{airshipFormation.pdf}}%
  \end{picture}%
\endgroup%

	\caption{Basic airship configurations in a \emph{leader/follower} formation.}
	\label{fig:airshipFormation}
\end{figure}

The idea of the ``leader-follower'' technique is to impose speed and orientation references for two airships at a time, which may fly in 3 different modes: in column, in line or in ``oblique'' formation (\cref{fig:airshipFormation}). The distance and orientation between the two airships are controlled by keeping the heading and relative speed between both, which is guaranteed by the low level controllers. This structure is then combined to generate more complex formations with an arbitrary number of airships, such as the known ``V'' formation. Although scalable to more UAVs, in this preliminary work, we consider only three airships in a ``V'' formation and seven airships in a ``hexagon'' formation. In addition, we extend the idea of \cite{bicho2006airship} to cover more complex flight mission cases like hovering and moving target tracking. 

\begin{figure}[!htb]
	\centering
	\def\svgscale{.7}
\begingroup%
  \makeatletter%
  \providecommand\color[2][]{%
    \errmessage{(Inkscape) Color is used for the text in Inkscape, but the package 'color.sty' is not loaded}%
    \renewcommand\color[2][]{}%
  }%
  \providecommand\transparent[1]{%
    \errmessage{(Inkscape) Transparency is used (non-zero) for the text in Inkscape, but the package 'transparent.sty' is not loaded}%
    \renewcommand\transparent[1]{}%
  }%
  \providecommand\rotatebox[2]{#2}%
  \newcommand*\fsize{\dimexpr\f@size pt\relax}%
  \newcommand*\lineheight[1]{\fontsize{\fsize}{#1\fsize}\selectfont}%
  \ifx\svgwidth\undefined%
    \setlength{\unitlength}{128.50881021bp}%
    \ifx\svgscale\undefined%
      \relax%
    \else%
      \setlength{\unitlength}{\unitlength * \real{\svgscale}}%
    \fi%
  \else%
    \setlength{\unitlength}{\svgwidth}%
  \fi%
  \global\let\svgwidth\undefined%
  \global\let\svgscale\undefined%
  \makeatother%
  \begin{picture}(1,1.43376933)%
    \lineheight{1}%
    \setlength\tabcolsep{0pt}%
    \put(0,0){\includegraphics[width=\unitlength,page=1]{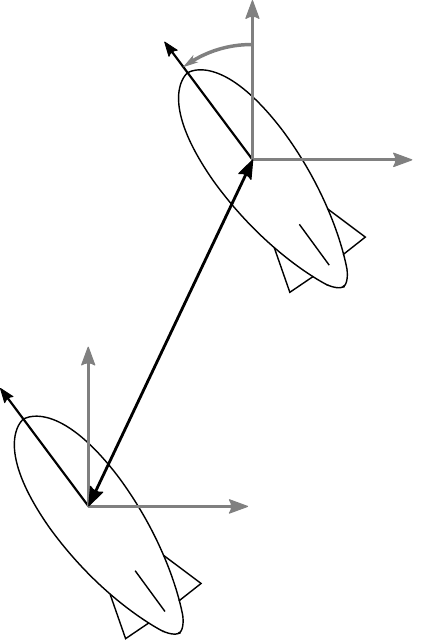}}%
    \put(0.37572949,1.3678303){\color[rgb]{0,0,0}\makebox(0,0)[lt]{\lineheight{1.25}\smash{\begin{tabular}[t]{l}$\psi_j$\end{tabular}}}}%
    \put(0.0255263,0.59399736){\color[rgb]{0,0,0}\makebox(0,0)[lt]{\lineheight{1.25}\smash{\begin{tabular}[t]{l}$\psi_i$\end{tabular}}}}%
    \put(0,0){\includegraphics[width=\unitlength,page=2]{controlErrors.pdf}}%
    \put(0.21540132,0.58394781){\color[rgb]{0,0,0}\makebox(0,0)[lt]{\lineheight{1.25}\smash{\begin{tabular}[t]{l}$\delta_i$\end{tabular}}}}%
    \put(0.40739914,0.65856315){\color[rgb]{0,0,0}\makebox(0,0)[lt]{\lineheight{1.25}\smash{\begin{tabular}[t]{l}$\rho_i$\end{tabular}}}}%
    \put(0,0){\includegraphics[width=\unitlength,page=3]{controlErrors.pdf}}%
    \put(0.58148242,1.34301641){\color[rgb]{0,0,0}\makebox(0,0)[lt]{\lineheight{1.25}\smash{\begin{tabular}[t]{l}$\epsilon_i$\end{tabular}}}}%
  \end{picture}%
\endgroup%

	\caption{Polar coordinate system with the origin set at the NED frame of the leader airship, used in the 2nd layer (SFKC), here with $ \rho_{i,d} = \zeta_{i,d} = 0. $}
	\label{fig:controlErrors}
\end{figure}

Consider that $\psi_i$ is the Euler angle orientation (yaw angle), in the D axis ($\mathfrak{I}$ frame) for the \textit{follower} airship and $\zeta_i$ the angle between the \textit{leader} and \textit{follower} positions in the same axis. It is desired that the \textit{follower} airship maintains a certain distance, $\rho_{i,d}$, from the \textit{leader}, as well as a certain relative angle, $\zeta_{i,d} $. The solution for this kind of problem of position/orientation control is delegated to the low level controller (3rd layer of \cref{fig:architecture}) that is based on a simple kinematic model (\cref{eq:modCarro} and \cref{fig:controlErrors}). This {SFKC} approach is commonly used in mobile robotics for terrestrial vehicles \cite{siegwart2011introduction}. 
 
\begin{equation}
	\begin{bmatrix}
		\dot{\rho}     \\
		\dot{\zeta}    \\
		\dot{\epsilon}
	\end{bmatrix}
	=
	\begin{bmatrix}
		-\cos{\zeta}              & 0  \\
		\frac{\sin{\zeta}}{\rho}  & -1 \\
		-\frac{\sin{\zeta}}{\rho} & 0 
	\end{bmatrix}
	\begin{bmatrix}
		u_{ref}      \\
		r_{ref}
	\end{bmatrix}
	\label{eq:modCarro}
\end{equation} 

It is important to remark that \cref{eq:modCarro} assumed the goal at the origin of the inertial frame, which does not incur any loss of generality. The complete relations are then given by:
\begin{equation}\label{eq:rho}
	\rho_i = \sqrt{{\Delta P_E}^2 + {\Delta P_N}^2} - \rho_{i,d}
\end{equation}	
\begin{equation}\label{eq:delta}
	\zeta_i = - \psi_i + \atan \frac{\Delta P_E}{\Delta P_N} - \zeta_{i,d} 
\end{equation}	
\begin{equation}\label{eq:epsilon}
	\epsilon_i = - \psi_i - \zeta_i + \psi_{j}
\end{equation}
which takes into consideration that the goal frame is a moving one. To derive the state feedback gain, we first linearise \cref{eq:modCarro} assuming small-angle errors such that $\cos(\zeta) \simeq 1$ and $\sin(\zeta) \simeq \zeta$. The feedback control law defined by \cite{siegwart2011introduction} is:
\begin{equation}
	u_{ref} = k_\rho \rho  
\end{equation}
\begin{equation}
	r_{ref} = k_{\zeta} \zeta + k_{\epsilon} \epsilon
\end{equation}
which had the subscripts dropped, in order to ease the notation. These two velocities, that correspond to the longitudinal linear speed and the yaw rate angular speed, compose the airspeed reference command vector sent to the airship as control inputs:
\begin{equation}\label{eq:SFKCoutput}
	\vect{u}_{a_{ref}}=[u_{ref} \ 0 \ 0 \ 0 \ 0 \ r_{ref}]\T;
\end{equation}
As the use of airspeed control signals infers no loss in generality, the subscript $ a $ will not be employed to ease the notation.

The substitution of these control laws in the linearised model of \cref{eq:modCarro} leads to the following closed-loop state space model:
\begin{equation}\label{eq:modCinCarro}
	\begin{bmatrix}
		\dot{\rho}     \\
		\dot{\zeta}    \\
		\dot{\epsilon}
	\end{bmatrix}
	=
	\begin{bmatrix}
		- k_\rho & 0                   & 0            \\
		0        & -(k_\zeta - k_\rho) & - k_\epsilon \\
		0        & -k_\rho             & 0
	\end{bmatrix}
	\begin{bmatrix}
		\rho   \\
		\zeta \\
		\epsilon
	\end{bmatrix}
\end{equation}
whose dynamic matrix should have stable eigenvalues by choosing $ k_\rho > 0 $, $ k_\epsilon < 0 $ and $ k_\zeta > k_\rho $.

An important adaptation made in the {SFKC} was the addition of a feedforward term to improve the velocity tracking of the follower airship. The velocity reference is set as the true airspeed of the leader airship $ V_t  = \norm{\vect{v}_a}_2 = \sqrt{u_a^2 + v_a^2 + w_a^2} $. If we consider $ y_j(k) = [P_{N_j}(k), \ P_{E_j}(k)]\T $ as the position of the leader airship at any instant $ k $, then the equilibrium point sought by the follower airship is given by 
\begin{equation}
\begin{split}
    y_{ref}(k) = 
    \begin{bmatrix}
        P_{N_j}(k) - \rho_{i,d} \cos({\psi_j(k) + \zeta_{i,d}}) \\
        P_{E_j}(k) - \rho_{i,d} \sin({\psi_j(k) + \zeta_{i,d}})
    \end{bmatrix}
\end{split}
\end{equation}
and 
\begin{equation}
	V_{t_{ref}} = \frac{y_{ref}(k) - y_{ref}(k-1)}{T_s}
\end{equation}
where $ T_s $ is the sampling time. This term corresponds to a feedforward gain $ k_{ff} $  in the closed-loop state model of the controller (as $ u_{ref} \approx V_{t_{ref}}$), or:
\begin{equation}\label{eq:modWaypoint}
\begin{split}
	\begin{bmatrix}
		\dot{\rho}     \\
		\dot{\zeta}    \\
		\dot{\epsilon}
	\end{bmatrix}
	=
	\begin{bmatrix}
		- k_\rho & 0                   & 0            \\
		0        & -(k_\zeta - k_\rho) & - k_\epsilon \\
		0        & -k_\rho             & 0
	\end{bmatrix}
	\begin{bmatrix}
		\rho     \\
		\zeta    \\
		\epsilon
	\end{bmatrix} + \\
	\begin{bmatrix}
		k_{ff} & 0 \\
		0      & 0 \\
		0      & 0
	\end{bmatrix} 
	\begin{bmatrix}
		u_{ref} \\
		r_{ref}
	\end{bmatrix}
\end{split}
\end{equation}

One of the most interesting properties of this adapted {SFKC} is that, instead of \cite{bicho2006airship}, with a single set of tuning control parameters $k_\zeta$, $ k_\rho $, $k_\epsilon$ and $ k_{ff} $ it is possible to switch, with just some small changes, between the different operational modes of waypoint flight, hovering flight and ground tracking flight used to execute the different airship missions. For the ``follower'' airship in the ``leader/follower'' guidance approach, the ``goal'' point in \cref{fig:controlErrors} will always be the position of its corresponding ``leader'' airship. And for the main leader airship in the ``V'' coordinated flight, the goal may be the next waypoint (in the waypoint tracking task), the hovering waypoint (in the hovering flight) or the target current position in North$ \times $East coordinates (in the target tracking). In this last case, the idea is to generate a kind of ``walk-stop-walk'' behaviour to follow the target using both the constant airspeed and the hovering control modes alternately. The airship will then follow the ground target at a constant airspeed, and whenever its projected distance to the object reaches a chosen limit, it switches to the hovering control mode. 

\vspace{-0.2cm}
\subsection{Third Layer - Sliding Mode Controller}
For the third (low level) layer of this control architecture, many different linear and nonlinear controllers have already been proposed and implemented for the AURORA and Noamay airships \cite{Paiva2006}, \cite{moutinho2007modeling}. However, when the airship has to switch between different mission tasks from low to high airspeeds, a robust nonlinear controller is usually required. That is the reason why we have chosen here the Sliding Mode Control (SMC) aproach implemented by \cite{Vieira2017}, as it manages to control the coupled airship system (lateral and longitudinal motions) while achieving great robustness, even under saturation of the actuators and for a wide range of airspeeds \SIrange[per-mode=symbol]{3}{13}{\meter\per\second}. 
In the following we just point out the  adaptations made in this control approach to cope with the proposed challenges and tasks of this work. The core design development can be seen in \cite{Vieira2017}.

As usual, the SMC controller has its gains calculated around certain equilibrium conditions, which means that it really only has peak performance around this condition, despite guaranteeing global stability. The dynamical conditions of the airship, however, are extremely non linear. There are saturations and dynamics on the actuators; a naturally elevated inertia, due to the airship form factor; high lateral wind drag; added mass due to air displaced; underactuation on the side of the airship and many more complications. That is to say that, despite being able to reject any uncertainty or perturbations, the gains calculated for a certain trim condition can have poor performance on different conditions, sometimes even saturating the controller. For that reason, it was decided to utilise a minimisation function, aiming to select the most appropriate gains at each instant of simulation. The following algorithm explains the minimisation function:
\begin{algorithm}
	\caption{trimMinimization($\vect{u}_{a_{ref}},\vect{V}_{t_e},\mathrm{SMCgains}$)}
	\begin{algorithmic}[1]
		\STATE $\vect{V}_{t_{ref}} \gets \vect{u}_{a_{ref}} - \vect{V}_{t_e}$ \% Vector $ \vect{V}_{t_e} $ will be subtracted of scalar $ u_{a_u} $
		\STATE $[\sim,I] \gets \min(\abs{\vect{V}_{t_{ref}}})$  \% The absolute value \\ of each element of $ \vect{V}_{t_{ref}} $ is calculated and \\ the index of the smallest is saved
		\STATE $k_{SMC} \gets \mathrm{SMCgains}(I)$ \% The gains of \\ the SMC for that trim condition are \\ selected and saved
		\RETURN $k_{SMC}$
	\end{algorithmic}
\end{algorithm}\\
where $\vect{u}_{a_{ref}}$ is the SFKC guidance controller output to the SMC controller (\cref{eq:SFKCoutput}), $\vect{V}_{t_e}$ is a vector that contains 74 different airspeeds - between [0.3-15] m/s - that are used as trim points, $ \mathrm{SMCgains}$ is another vector that contains 74 sets of gains for the SMC controller, related to each trim point and $\vect{V}_{t_{ref}}$ is the reference airspeed. In other words, at each instant the function will choose the closest trim point to the desired airspeed and will send the corresponding set of gains to the SMC controller.

\section{Swarm Intelligence}
\label{sec:intelligentswarm}

\vspace{+10pt}

We present below two swarm intelligence strategies used in this work: the Boids Algorithm \cite{reynolds1987flocks} and the Robotic Particle Swarm Optimization (RPSO) \cite{couceiro2011novel}.

\subsection{Boids Algorithm}

The Boids algorithm has first been designed by Craig Reynolds in \cite{reynolds1987flocks}, in order to simulate a flock of birds (or Boids) in a more realistic way. It is based on three simple rules:

\begin{enumerate}[label=(\roman*)]
\item \textit{Collision Avoidance}: the members of the flock are not supposed to collide among
themselves (short range \textit{repulsion}).
\item \textit{Velocity Matching}: the members align their speed and direction on their neighbours’ ones (neighbours \textit{mimicking}).
\item \textit{Flock Centring}: the members have to stay together to form a group (long range
\textit{attraction}).
\end{enumerate}

\vspace{+5pt}
The algorithm can be summarised as follows. Let us consider a flock of $N$ Boids, which $i^{th}$ Boids state is given by:

\begin{equation}
\vect{x}_i = (\vect{p}_i^T,\vect{v}_i^T)^T 
\end{equation}
where $\vect{p}_i$ is the Boid's position and $\vect{v}_i$ its velocity with respect to a fixed global frame. The state of a Boid  at time step $t$ can be written as:

\begin{equation}
\vect{x}_{i,t} = \vect{x}_{i,t-1} + \vect{v}_{i,t} \Delta_t
\end{equation}
where $\vect{v}_{i,t}$ is the velocity of the $i^{th}$ member of the swarm at time $t$, and $\Delta_t=t_i-t_{i-1}$ is the time increment. 

The three rules, \textit{collision avoidance/repulsion}, \textit{velocity matching/mimicking} and  \textit{flock centring/attraction} contribute to update each agent's velocity $\vect{v}_i$ at every time step $t$. The update can be expressed by weighted average as:

\begin{equation}
\vect{v}_{i,t}=\delta(\vect{v}_{i,t-1})+(1-\delta)(k_r.\vect{v}_{i,r}+k_m.\vect{v}_{i,m}+k_a.\vect{v}_{i,a})
\end{equation}
where $\vect{v}_{i,r}$, $\vect{v}_{i,m}$ and $\vect{v}_{i,a}$ represent respectively the computation of the 3 aforementioned rules, \textit{repulsion, mimicking, attraction} and $k_r$, $k_m$ and $k_a$, are the respective weights which enable the designer to tune the behaviour of the flock. 

\noindent The velocities $\vect{v}_{i,r}$, $\vect{v}_{i,m}$ and $\vect{v}_{i,a}$ are defined as follows:

\vspace{+5pt}
Velocity $\vect{v}_{i,r}$:

\begin{equation}
\vect{v}_{i,r}=\sum_{j\in \mathcal{E}_i}{\frac{\vect{p}_{i,t-1}-\vect{p}_{j,t-1}}{N_i'-1}}
\label{eq:rep}
\end{equation}

\noindent where $\vect{p}_{i,t-1}$ is the position of the $i^{th}$ agent at time $t-1$, and $\mathcal{E}_i$ and $N_i'$ are defined as:

\begin{equation}
\mathcal{E}_i = \{\ell \in [1, N]  \subset \mathbb{N} ~|~ \lVert \vect{p}_{i,t-1}-\vect{p}_{\ell,t-1} \rVert \leq d_{lim}\}
\label{eq:set}
\end{equation}

\begin{equation}
N_i' = card \lbrace \mathcal{E}_i \rbrace
\end{equation}

\noindent and $d_{lim}$ sets threshold distance, below which the agents will avoid each other.

\vspace{+5pt}
Velocity $\vect{v}_{i,m}$: 
\begin{equation}
\vect{v}_{i,m}=\sum_{j=1,j\neq i}^{N}{\frac{\vect{v}_{j,t-1}}{N-1}}
\label{eq:mim}
\end{equation}

\vspace{+5pt}
Velocity $\vect{v}_{i,a}$:
\begin{equation}
\vect{v}_{i,a}=-\vect{p}_{i,t-1} + \sum_{j=1,j\neq i}^{N}{\frac{\vect{p}_{j,t-1}}{N-1}}
\label{eq:att}
\end{equation}

As can be seen in (\ref{eq:set}), a limit distance $d_{lim}$ corresponds to the distance from which the separation rule starts having an effect on the Boid. Indeed, this separation rule only have a short range effect. The behaviours described by (\ref{eq:mim}) and (\ref{eq:att}) can be enhanced by adding other limit distances. This idea comes directly from natural flocks, where the members only perceive their local environment and closest neighbours, and do not have information from the whole group of animals. The Boids algorithm  has already been successfully implemented on a real robotics swarm, for example in \cite{saska2014swarms}.

\vspace{+10pt}
\subsection{Robotic Particle Swarm Optimization}

The second swarm intelligence based approach is the Robotic Particle Swarm Optimization (RPSO) proposed by Couceiro and collaborators in \cite{couceiro2011novel} which is based on the original PSO developed by Russel Eberhard and James Kennedy in \cite{eberhart1995new}. As every optimisation algorithm, it aims at finding the global optimum of a fitness function (also called cost function). It is based on a group of cooperating potential solutions (the particles) moving in the hyperspace of solutions. In other words, each particle can be considered as an explorer moving on a landscape of solutions, and communicating with its teammates to find the best one. It can be seen as a swarm control algorithm adapted to solve optimisation problems. In a robotics application, the particles are the robots themselves \cite{pessin2013swarm}, and the cost function depends on their distance to the target (in the case of a search mission for example). The robots are controlled by the model described in Equation \ref{eq:rpsoVel} and Equation \ref{eq:rpsoPos}, with operator $\otimes$ representing the tensor product of vectors.

\begin{equation}\label{eq:rpsoVel}
\begin{aligned}
\vect{v}_{t+1}^{ref}&=\vect{a} \otimes \vect{v}_t^{ref} \\
&+ \vect{b}_1 \otimes  r_1 \otimes  (\vect{p}_{pb,t} - \vect{x}_t)\\
&+ \vect{b}_2 \otimes  r_2 \otimes  (\vect{p}_{nb,t} - \vect{x}_t)\\
&+ \vect{b}_3 \otimes  r_3 \otimes  (\vect{p}_{obs,t} - \vect{x}_t)
\end{aligned}
\end{equation}
\begin{equation}
\vect{x}_{t+1} = \vect{x}_t + \vect{v}_{t+1} \Delta_t
\label{eq:rpsoPos}
\end{equation}
\noindent where $\mathbf{v}_{t+1}^{ref}$ is the updated velocity reference,  $\mathbf{v}_t$ is the current velocity vector of the robot (which does not correspond to the current reference velocity vector),
$\mathbf{p}_{pb,t}$ is the personal best position of the robot (i.e. the position where the robot
scored better) $\mathbf{p}_{nb,t}$ is the neighbours' best position (i.e. the best personal best
position among the swarm) and $\mathbf{p}_{obs,t}$ is a repulsion vector which minimises the risks of collisions with the other robots. Finally, $\mathbf{p}_t$ is the current position of the airship, used in the RPSO strategy detailed in Algorithm \ref{alg:rpso2d}. 


\begin{algorithm}[!htb]
	\caption{RPSO algorithm for a swarm of $N$ robots in a dynamic environment} 
	\label{alg:rpso2d} 
	\begin{algorithmic}
		\STATE \textbf{Inputs:} Matrix of the current positions of the robots
		$\mathbf{P}_t=\left[\mathbf{p}_{1,t}\cdots\mathbf{p}_{N,t}\right]$\\
		Matrix of the current velocities of the robots
		$\mathbf{V}_t=\left[\mathbf{v}_{1,t}\cdots\mathbf{v}_{N,t}\right]$\\
		Matrix of the personal best positions of the robots
		$\mathbf{P}_{pb,t-1}=\left[\mathbf{p}_{pb,1,t-1}\cdots\mathbf{p}_{pb,1,t-1}\right]$\\
		Vector of the corresponding fitnesses $\mathbf{f}_{pb,t-1}=\left(f_{pb,1,t-1}\cdots
		f_{pb,N,t-1}\right)$
		\STATE \textbf{Output:} Matrix of the velocity vectors for the robots of the swarm
		$\mathbf{V}_{t+1}^{ref}=\left[\mathbf{v}_{1,t+1}^{ref}\cdots\mathbf{v}_{N,t+1}^{ref}\right]$
		\STATE
		\FORALL{Robot $i$ in the swarm}
			\STATE Recompute the fitness $f_{pb,i,t}$ of the personal best position of the robot
			$\mathbf{p}_{pb,i,t-1}$
			\STATE $f_{pb,i,t-1} \gets f_{pb,i,t}$
			\STATE Compute the new fitness $f_{i,t}$ of the robot at position $\mathbf{p}_{i,t}$
			\IF{$f_{i,t}\geq f_{pb,i,t}$}
				\STATE $f_{pb,i,t}\gets f_{i,t}$
				\STATE $\mathbf{p}_{pb,i,t}\gets \mathbf{p}_{i,t}$
			\ENDIF
		\ENDFOR
		\FORALL{Robot $i$ in the swarm}
			\STATE Compute $\mathbf{v}_{rep,i,t}$ with eq(\ref{eq:rep})
			\STATE Get $\mathbf{v}_{i,t}$ from $\mathbf{V}_t$
			\STATE Get $\mathbf{p}_{pb,i,t}$ from $\mathbf{P}_{pb,t}$
			\STATE Initialise $\mathbf{p}_{nb,t}=\mathbf{0}$ and $f_{nb,t}=0$
			\FORALL{$f_{i,k}$ in $\mathbf{f}_t$}
				\IF{$f_{i,k}>f_{nb,t}$}
					\STATE $f_{nb,t}\gets f_{i,t}$
					\STATE $\mathbf{p}_{nb,t} \gets \mathbf{p}_{pb,i,t}$
				\ENDIF
			\ENDFOR
			\STATE Compute $\mathbf{v}_{i,t+1}^{ref}$ with eq(\ref{eq:rpsoVel})
		\ENDFOR
		\RETURN $\mathbf{V}_{t+1}^{ref}$
	\end{algorithmic}
\end{algorithm}

\begin{figure}[!htb]
    \centering
    \includegraphics[width=\columnwidth]{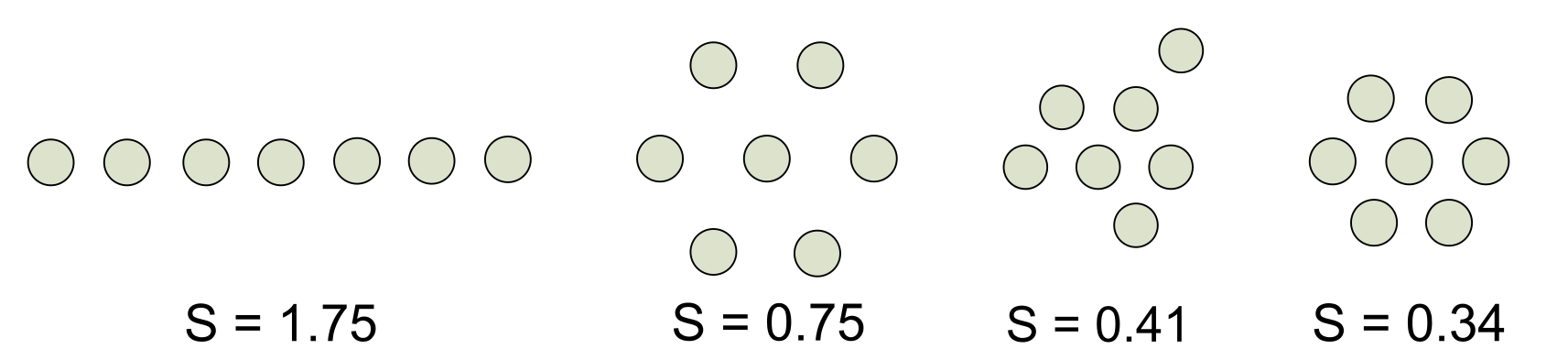}
    \caption{Entropy of different robot formations, from \cite{turgut2008self}}
    \label{fig:entropy_ex}
\end{figure}

 In this particular work, two measures are taken into account in the fitness function ($\mathbf{f}_{pb,t}$) that is optimized: distance to waypoint/target and \textit{swarm's entropy}. Entropy ($S$) in swarms was defined in\cite{turgut2008self}, and measures the \textit{disorganization} of the swarm, position-wise. Figure \ref{fig:entropy_ex} exemplifies the entropy values in different robot configurations.

\begin{figure}
    \centering
    \includegraphics[width=0.5\columnwidth]{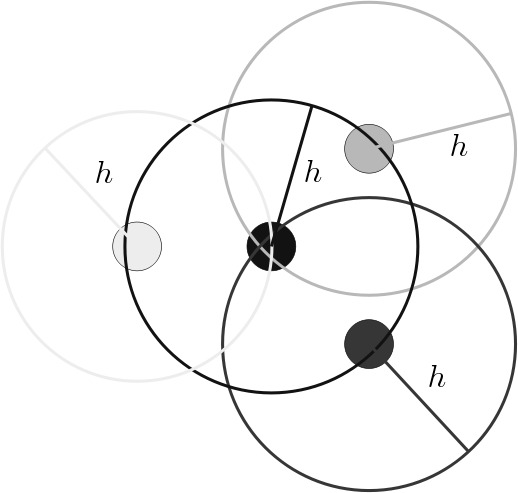}
    \caption{Example of clusters radii.}
    \label{fig:cluster}
\end{figure}

\noindent To compute (S) a cluster-based technique is employed, and studies the variation of agents in a cluster, by varying a cluster's radius $h$ centered around each robot. Figure \ref{fig:cluster} exemplifies the membership of agents to a set of clusters.  

\noindent In sum, two agents are in the same cluster if and only if  $||\mathbf{p}_i-\mathbf{p}_j||\leq h$ where $\mathbf{p}_i$ and $\mathbf{p}_j$ represent the position vectors of the $i^{th}$ and $j^{th}$ airships.
To compute entropy, a formal definition of social entropy, given by Bach, in \cite{balch2000hierarchic} is used:
\begin{equation}
H(h)=\sum_{k=1}^M p_k\, log_2(p_k)
\label{eq:entropy}
\end{equation}
where $p_k$ is the ratio between the individuals in the $k^{th}$ cluster and the total number of individuals in the swarm, and $M$ is the number of clusters for a given $h$. These values are integrated over $h$ ranging from 0 to $\infty$ to find total entropy ($S$) of the swarm.
\begin{equation}
S=\int_0^{\infty} H(h) ~dh
\label{eq:entropy_total}
\end{equation}
An algorithm to compute $(S)$ is provided in Appendix \ref{app:entropy}.\\
The second component of the fitness function is simply distance to target defined as:
\begin{equation}
    d_{i_t} = \|\mathbf{p}_{i,t}-\mathbf{p}_{tar,t}\|
    \label{eq:dist_to_target}
\end{equation}
Therefore the fitness of robot $i$ at time $t$, ($f_{i,t}$) will be a weighted sum between functions of entropy ($S$) and distance to waypoint/target ($d_{i,t}$) given by (\ref{eq:costfun}). 
The idea is that by moving according to (\ref{eq:rpsoVel}), the robot will optimise its fitness, and thus minimise
both the \emph{entropy} of the swarm and its own distance to the waypoint.
 
\begin{equation}
\begin{split}
f_{i,t}&=\gamma_S^2+\gamma_d^2\\
\gamma_S &= k_S \exp\left(\frac{-S^2}{2*R_S^2}\right)\\
\gamma_d &= k_d \exp\left(\frac{-d_{i,t}^2}{2*R_d^2}\right)\\
\label{eq:costfun}
\end{split}
\end{equation}
where:\\

\begin{tabular}{lll}
     $f_{i,t}$ &- & Value of the objective function for agent $i$\\
                &  & at time $t$. \\
     $\gamma_S$ and $\gamma_d$& - & Weighted functions of the two objectives:\\
     & & minimize entropy ($\gamma_S$) and minimize \\
     & & distance to waypoint ($\gamma_d$).\\
\end{tabular}

\begin{tabular}{lll}
     $k_S$ and $k_d$& - & Adjustable weights of the relative \\
                    &   & contribution of each objective for\\
                    &   & the value of $f_{i,t}$.\\
     $R_S$ and $R_d$& - & Adjustable parameters of the Gaussian\\
                    &   & functions.
\end{tabular}


\noindent As two variables have to be optimised at the same time ($\gamma_S$ and $\gamma_d$),
the cost function is the sum of these
squared parameters. The two
intermediate variables $\gamma_S$ and $\gamma_d$ correspond to a Gaussian function of $S$ and
$d_{i,k}$. They have two purposes:
\begin{itemize}
\item Normalise the value of the distance and of the \emph{entropy}, which
would otherwise be too different to be summed.
\item Transform these values into positive numbers that increase when the fitness
of the robot is better (closer to the target and more compact swarm).
\end{itemize}

\noindent The parameters $k_S$ and $k_d$ can be used to adjust the behaviour of the swarm: increasing
one will make the effect of the associated component of the fitness function more important compared to the other. Parameters $R_S$ and $R_d$ can be used to tune
the sensitivity of the Gaussian function, i.e. to adjust the width of the “bell”. If the
latter is too small or too big, then a small displacement of the robot may not change
significantly the fitness value and result in an undesirable behaviour. 






\section{Simulation Results}
\label{sec:simulations}
We present here the simulation results of the application of both cooperative flight controllers (formation flight and swarm intelligence techniques) for the two tasks (\textit{waypoint path following} and \textit{moving target tracking}). 

\subsection{Waypoint Path Following Task}
\subsubsection{Formation Flight}
\begin{figure}[!htb]
	\centering
	\includegraphics[width=\linewidth]{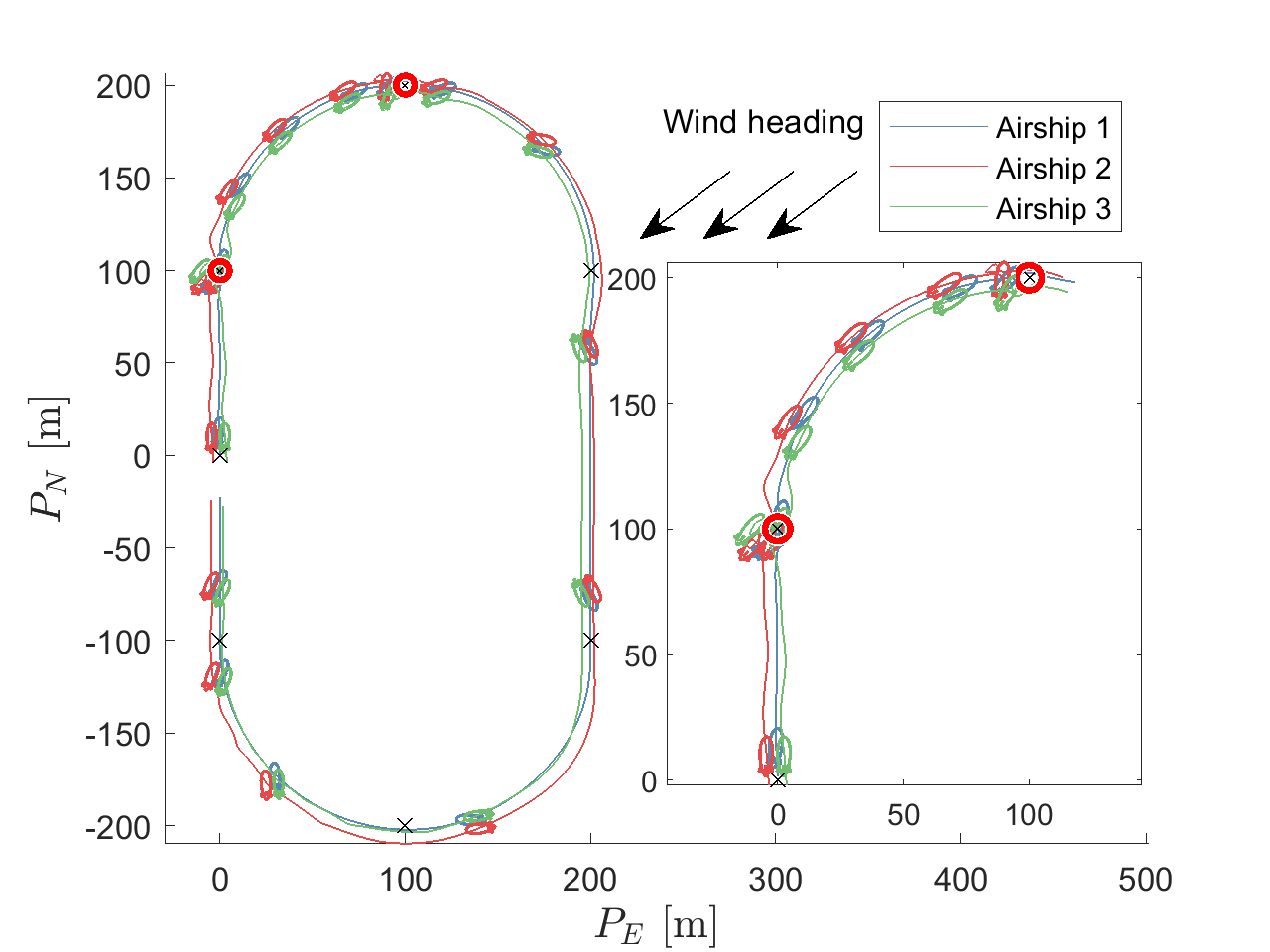}
	\caption{Airships in Waypoint/Hovering mode (and zoom). The red circles are the hovering areas. The airships are drawn at every 20 seconds.}
	\label{fig:waypointDynamicalHover}
\end{figure}

\begin{figure}[!htb]
	\centering
	\includegraphics[width=.85\linewidth]{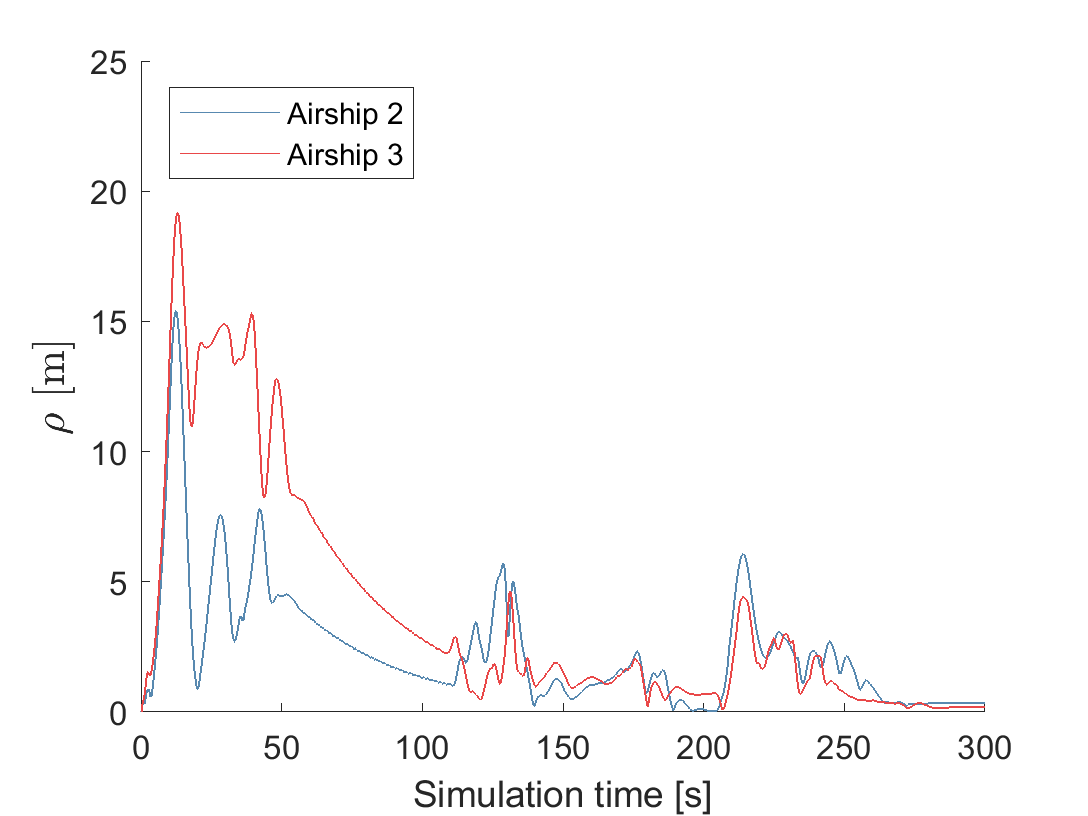}
	\caption{Euclidean norm of error plot of the follower airships in Waypoint/Hovering mission (\cref{fig:waypointDynamicalHover}). The first peak value at $t\approx\SI{20}{\second}$ correspond to the initial positioning and first hover point and at $t\approx\SI{125}{\second}$, to the second. The third peak is due to the \SI{180}{\degree} change in direction. The average error was \SI{3.05}{\meter} and the standard deviation \SI{3.26}{\meter}}
	\label{fig:waypointDynamicalHoverErrors}
\end{figure}

\begin{figure}[!htb]
	\centering
	\includegraphics[width=.85\linewidth]{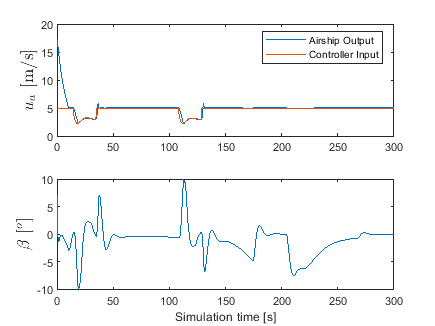}
	\caption{Leader's linear velocity ($u_a$, in blue) and control signal (in red), and sideslip angle ($ \beta $) during Waypoint/Hover mission (\cref{fig:waypointDynamicalHover}).}
	\label{fig:waypointDynamicalHoverSpeed}
\end{figure}

~\\
\Cref{fig:waypointDynamicalHover} shows the simulation of the waypoint navigation, for three airships, including eventual hovering points (red circles), with chosen limit radius of \SI{20}{\metre}. The mission duration was \SI{300}{\second}. The follower airships are able to trace a straight path that represented the instantaneous velocity of the leader airship, which allowed them to keep track of their lateral errors. As the leader tends to exhibit lateral errors during flight, the follower airships amplify these errors during the mission, i.e. their flight stability depends directly on the leader stability. To dampen this effect, they see their given formation point as a hover point of radius \SI{1}{\meter}, in this mission. Note that, over the hovering points, the airships align themselves against the wind when trying to stay in circle. In this simulation, the follower airships had an average of position error of \SI{3.05}{\meter} and a standard deviation of \SI{3.26}{\meter}, both calculated by the Euclidean norm of the difference. The errors are shown in \cref{fig:waypointDynamicalHoverErrors}.
\Cref{fig:waypointDynamicalHoverSpeed} shows the leader controller inputs and outputs signals for Airship 1. It is possible to see that the commanded airspeed in $ u_a $ is well tracked.
~\\
\subsubsection{Swarm Intelligence}
\begin{figure}[H]
    \centering
    \hspace{-10pt}
    \begin{subfigure}[b]{0.4\columnwidth}
        \includegraphics[height=3in]{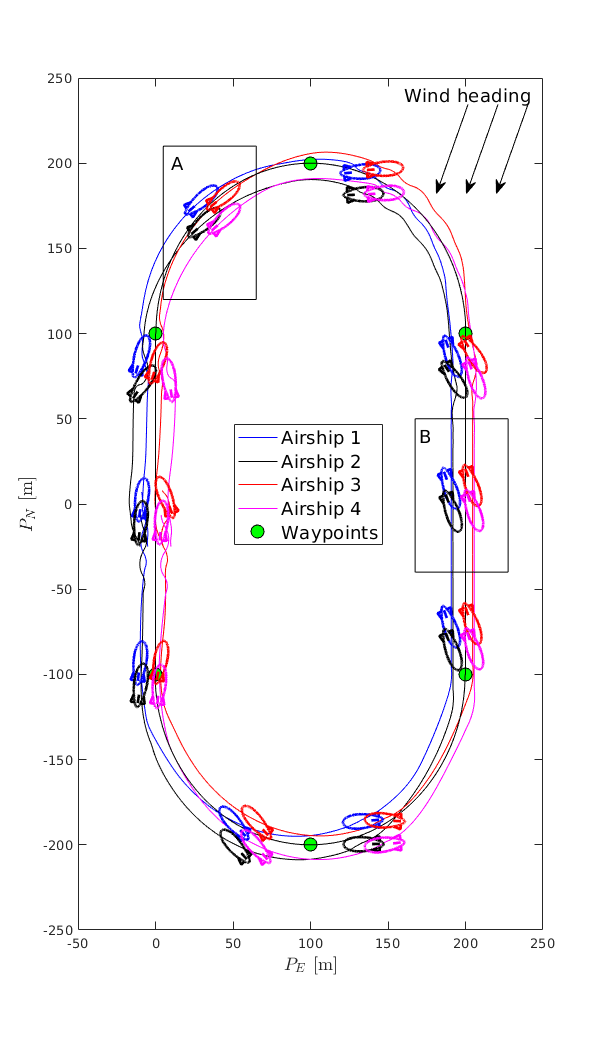}
        \caption{BOIDS approach}
        \label{subfig:boids}
    \end{subfigure}
    \hspace{20pt}
    \begin{subfigure}[b]{0.4\columnwidth}
        \includegraphics[height=3in]{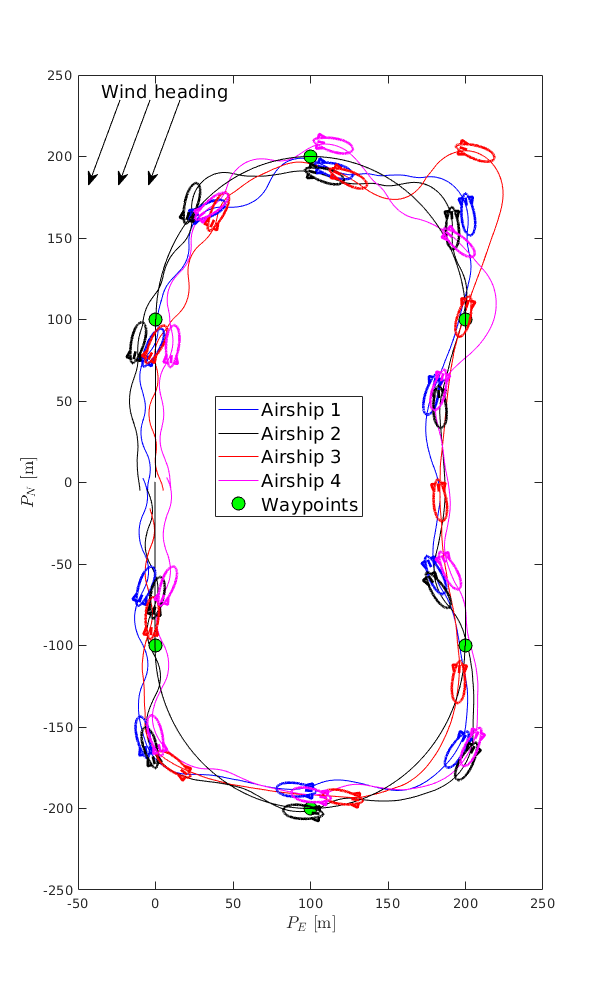}
        \caption{RPSO approach}
        \label{subfig:rpso}
    \end{subfigure}
    \caption{Swarm Intelligence approaches to Waypoint navigation}
    \label{fig:waypointSwarm}
\end{figure}

~\\
Below, we show the results for waypoint path following employing the swarm intelligence strategies described in section \ref{sec:intelligentswarm}. Figures \ref{subfig:boids} and \ref{subfig:rpso} show the resulting trajectories for four airships.\\
On one hand, Fig \ref{subfig:boids} shows a much smoother trajectory of the entire group. In this particular case, the coordinates of each waypoint are added sequentially as a "fictional" member of the airship swarm.

\begin{figure}[!htb]
    \centering
      \begin{subfigure}[b]{0.45\columnwidth}
        \centering
        \includegraphics[width=\textwidth]{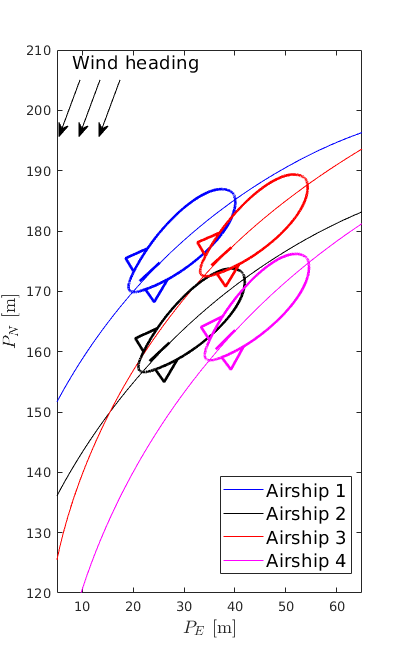}
        \caption{Detail A}
        \label{subfig:detA}
    \end{subfigure}
    \begin{subfigure}[b]{0.45\columnwidth}
        \centering
        \includegraphics[width=\textwidth]{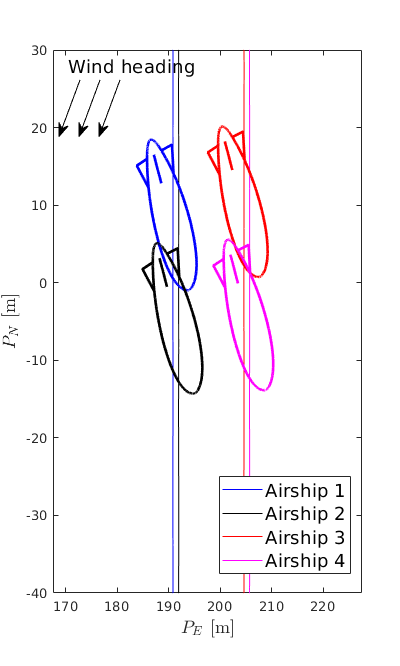}
        \caption{Detail B}
        \label{subfig:detB}
    \end{subfigure}
%
    \caption{Details of waypoint navigation using BOIDS.}
    \label{fig:details}
\end{figure}

 In practice, it means the waypoints' coordinates will contribute to the "Flock Centring" and "Collision Avoidance" rules, thus affecting the motion of the entire swarm. A particular aspect of this approach, in comparison to "formation flight" scenario, is that along the mission, the airships are able to change their relative position to each other, as highlighted in Figures  \ref{subfig:detA} and \ref{subfig:detB}.

The fact that the RPSO attempts to minimize a fitness function that is time dependent, i.e., the best distance between agents and to waypoint changes over time, is then the underlying reason for the almost sinusoidal behaviour of the swarm in Figure \ref{subfig:rpso}. 

\begin{figure}[!htb]
    \centering
      \begin{subfigure}[b]{1\columnwidth}
        \centering
        \includegraphics[width=.95\textwidth]{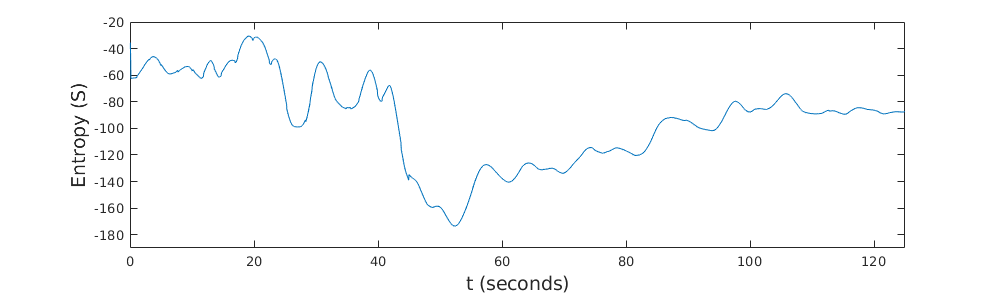}
        \caption{entropy(S) over time}
        \label{subfig:entropy}
    \end{subfigure}
    \begin{subfigure}[b]{1\columnwidth}
        \centering
        \includegraphics[width=.90\textwidth]{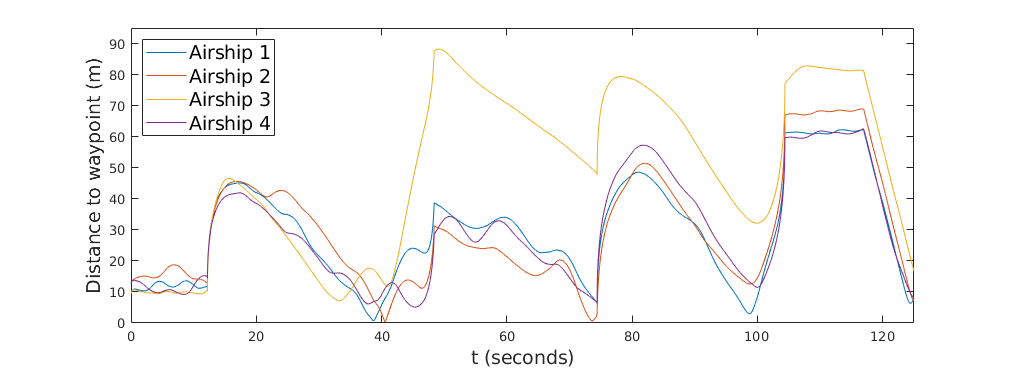}
        \caption{Distance of each Airship to waypoint}
        \label{subfig:dist}
    \end{subfigure}
    \caption{Metrics' evolution over time}
    \label{fig:metrics}
\end{figure}

Figures \ref{subfig:entropy} and \ref{subfig:dist} show the evolution over time of both Entropy, and the distance to the waypoint of each airship. It is important to highlight the steep changes in distance arrive from switching the current waypoint to the next one.
Furthermore, it is clear from these two figures that both objectives may not always be complementary, i.e., minimizing entropy may increase the distance to the waypoint, or decreasing this distance may increase Entropy. This trade-off is evident between $t=40s$ and $t=60s$, when decreasing entropy incurred in a larger distance of one of the airships to the waypoint.

\begin{figure}[!htb]
    \centering
    \hspace{-40pt}
    \begin{subfigure}[b]{0.4\columnwidth}
        \centering
        \includegraphics[height=3in]{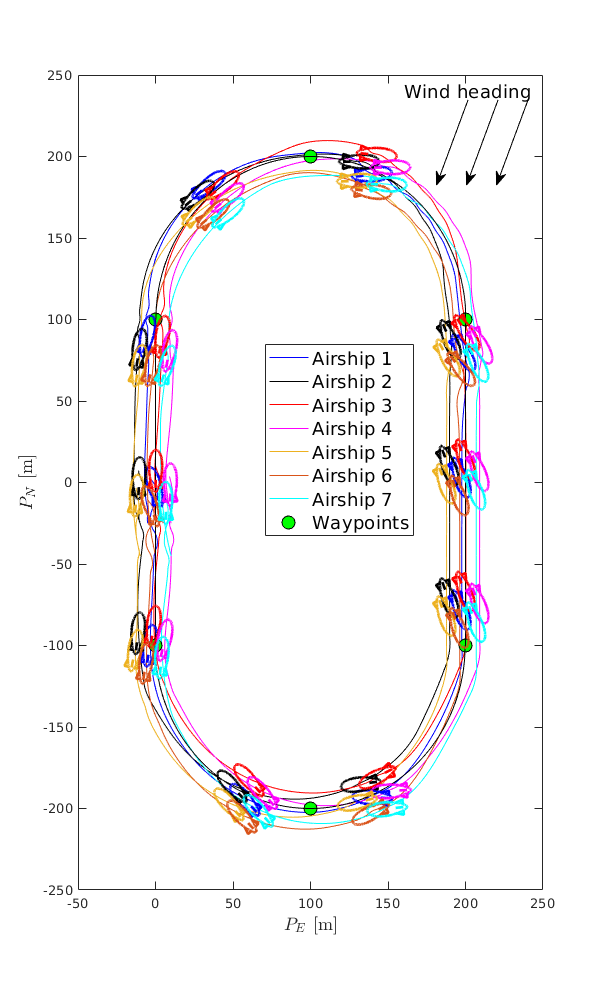}
        \caption{BOIDS approach}
        \label{subfig:boids7}
    \end{subfigure}
    \hspace{20pt}
    \begin{subfigure}[b]{0.4\columnwidth}
        \centering
        \includegraphics[height=3in]{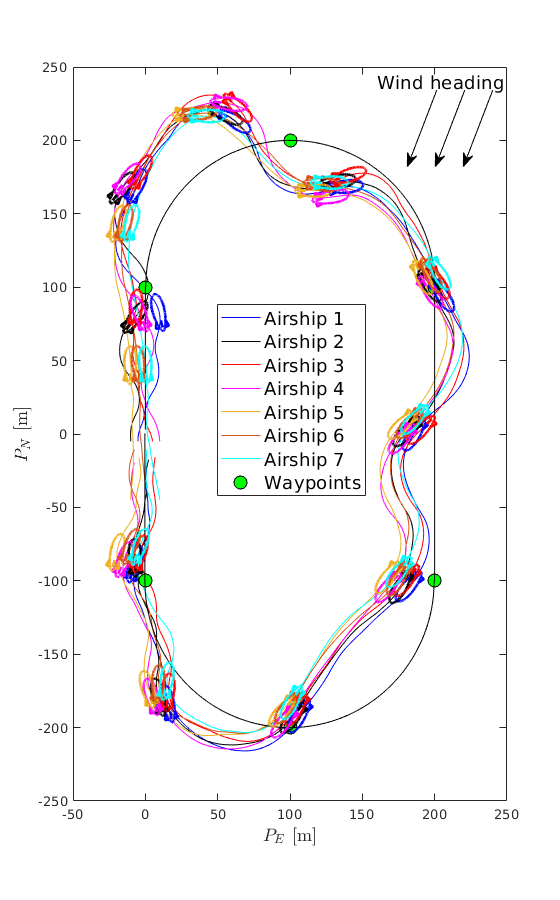}
        \caption{RPSO approach}
        \label{subfig:rpso7}
    \end{subfigure}
    \caption{Swarm Intelligence approaches to Waypoint navigation considering a group of 7 airships}
    \label{fig:waypointSwarm7}
\end{figure}

\subsection{Moving Target Tracking Task}
\label{sec:targettracking}

\subsubsection{Formation Flight}
\begin{figure}[!htb]
	\centering
	\includegraphics[width=\linewidth]{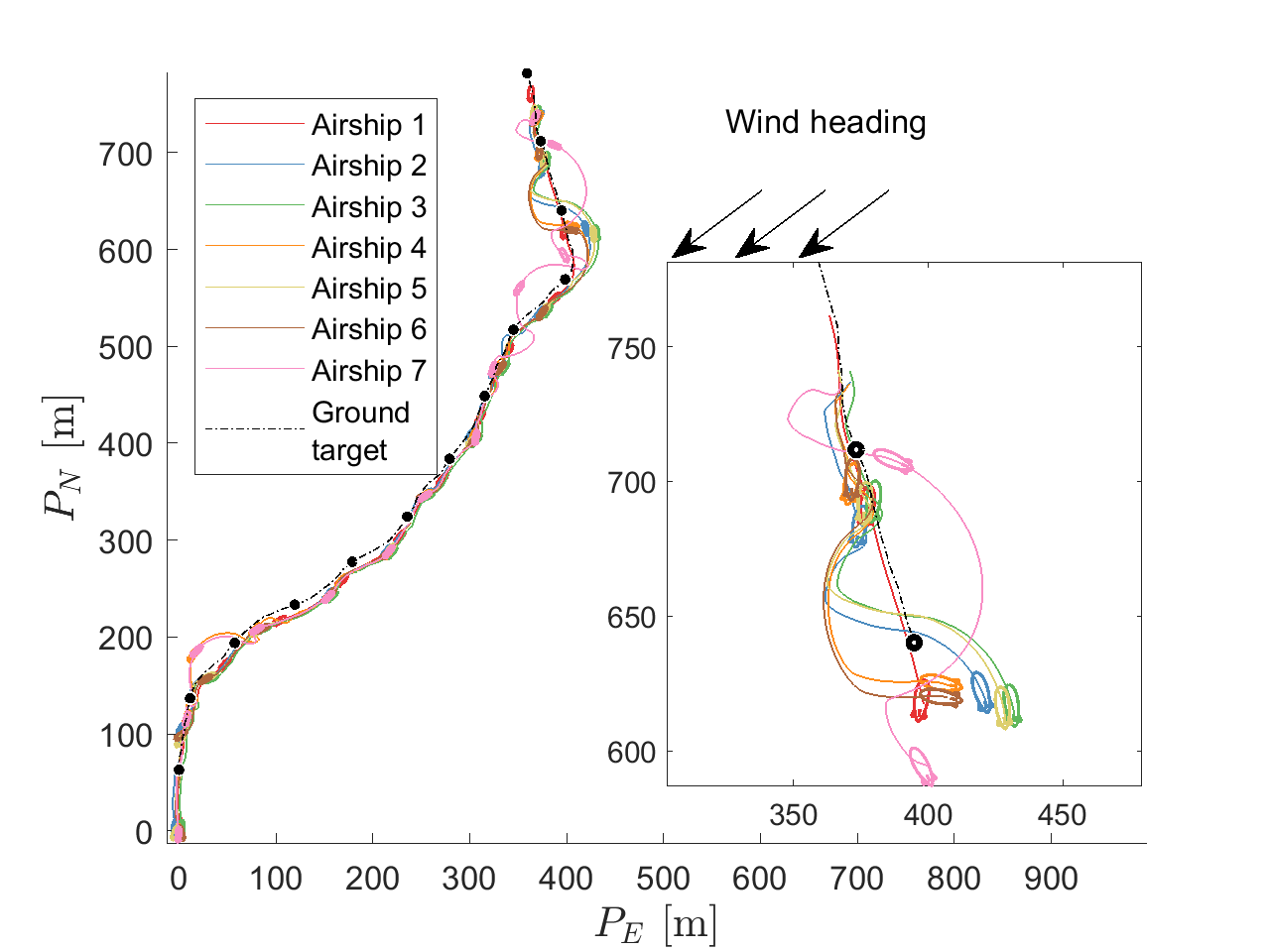}
	\caption{Airships in target tracking mission. The airships are drawn every 20 seconds.}
	\label{fig:waypointDynamicalTargetTracking}
\end{figure}

\begin{figure}[!htb]
	\centering
	\includegraphics[width=\linewidth]{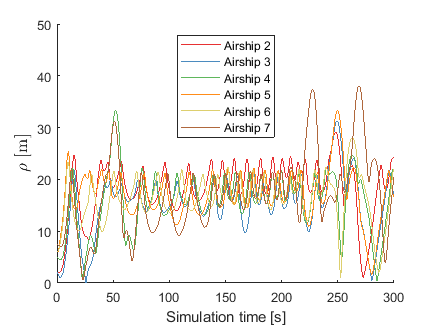}
	\caption{Euclidean norm of error plot of the follower airships in moving target mission (\cref{fig:waypointDynamicalTargetTracking}). The average error was \SI{16.66}{\meter} and the standard deviation \SI{3.78}{\meter}.
	\label{fig:waypointDynamicalTargetTrackingErrors}
	}
\end{figure}

\begin{figure}[!htb]
	\centering
	\includegraphics[width=\linewidth]{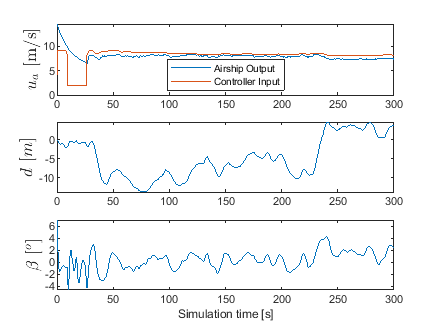}
	\caption{Leader's control signals and sideslip angle ($ \beta $) during target tracking (\cref{fig:waypointDynamicalTargetTracking}).}
	\label{fig:waypointDynamicalTargetTrackingSpeed}
\end{figure}

\begin{figure}[!htb]
	\centering
	\includegraphics[width=\linewidth]{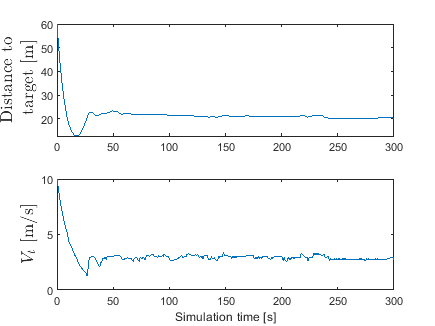}
	\caption{Leader's error and total velocity signals during target tracking (\cref{fig:waypointDynamicalTargetTracking}). The average error was \SI{21.30}{\meter} and the standard deviation \SI{3.75}{\meter}}
	\label{fig:waypointDynamicalTargetTrackingDynamics}
\end{figure}

~\\
The second simulation shown on \cref{fig:waypointDynamicalTargetTracking}, for 7 airships, illustrates the case of moving target tracking and \cref{fig:waypointDynamicalTargetTrackingSpeed} shows the corresponding leader's signals.
The error distribution from the followers is shown on \cref{fig:waypointDynamicalTargetTrackingErrors} and from the leader is shown on \cref{fig:waypointDynamicalTargetTrackingDynamics}.
In this mission, the leader is told to follow a ground target, by engaging successive hovering modes. In this case, the set radius for the hovering points is \SI{20}{\metre} for the leader and follower airships. In this case, increasing this radius acts a filter for the controller, avoiding overcompensation and erratic movement.

The leader had an average error of \SI{21.30}{\meter} and a standard deviation of \SI{3.75}{\meter}, while the follower airships had an average of position error of \SI{16.66}{\meter} and a standard deviation of \SI{3.78}{\meter}. It should be noted that a \SI{20}{\metre} is not compensated by the controller, due to the hoverpoint implementation.  

~\\
\subsubsection{Swarm Intelligence}
\begin{figure}[!htb]
    \centering
        \centering
        \begin{subfigure}[b]{\columnwidth}
        \centering
        \includegraphics[width=\columnwidth]{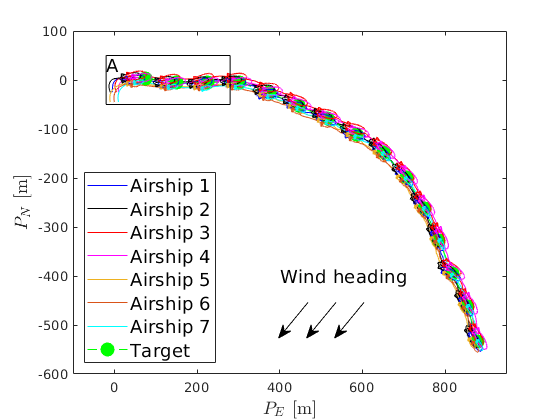}
        \caption{Full trajectory.}
        \end{subfigure}
        \\
        \begin{subfigure}[b]{\columnwidth}
        \centering
        \includegraphics[trim={80pt 0 80pt 0pt},clip,width=\columnwidth]{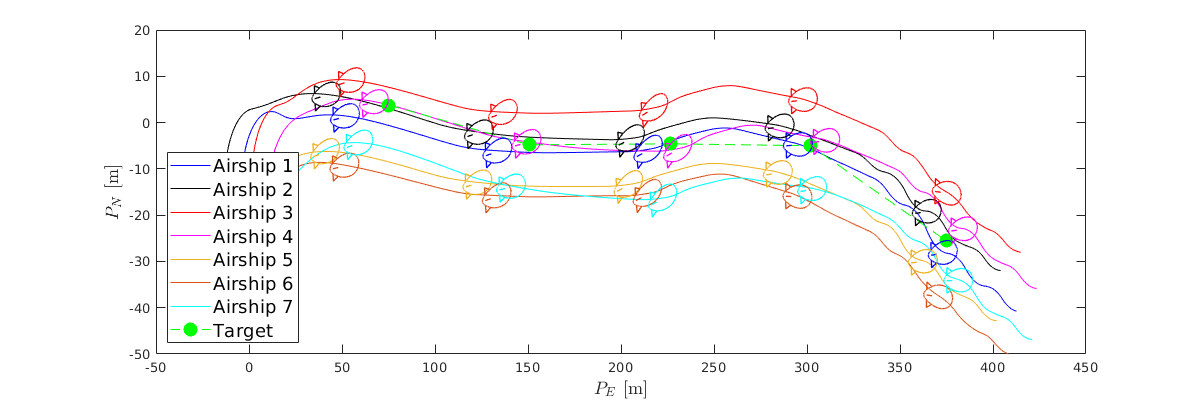}
        \caption{Detailed view}
        \end{subfigure}
        \caption{Trajectory of random moving target and a 7 airships swarm using BOIDS.}
        \label{fig:boidstrajectory7}
\end{figure}

\begin{figure}[b]
    \centering
        \centering
        \includegraphics[width=\columnwidth]{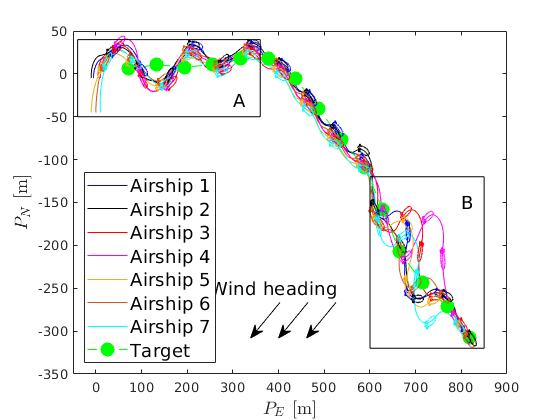}
        \caption{Full trajectory of random moving target and airship swarm using RPSO for 7 airships}
        \label{fig:rpsotrajectory7}
\end{figure}

~\\
This sub-section analyses the behaviour of the airship team considering the swarm intelligence strategies. Figures \ref{fig:boidstrajectory7} and \ref{fig:rpsotrajectory7} show, respectively, the full trajectory of both the BOIDS and the RPSO, plus a detailed view of the trajectories. Together, these figures validate the waypoint task, because when we make use of swarm intelligence strategies, a \textit{moving target tracking} task can be envisioned as a generic form of a \textit{waypoint path following} task.  For instance, the next waypoint path can move in real time. Since the rules of the swarm remain the same, the successful tracking of the moving target (or next waipoint path) shows the inherent flexibility of this swarm intelligence approach.%
%
%
%
%
%
\begin{figure}[!htb]
    \centering
        \begin{subfigure}[b]{\columnwidth}        
        \includegraphics[trim={20pt 0 20pt 20pt},clip,width=\columnwidth]{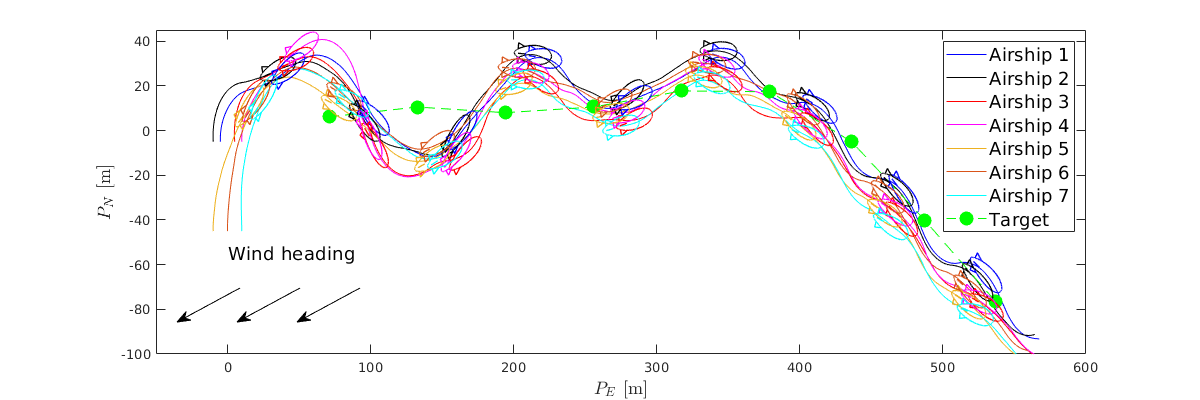}
        \caption{Detail A of the trajectory of random moving target and airship swarm using RPSO for 7 airships}
        \label{subfig:rpsodetailA7}
        \end{subfigure}
        \begin{subfigure}[b]{\columnwidth}
        \centering
        \includegraphics[trim={20pt 0 20pt 20pt},clip,scale=0.4]{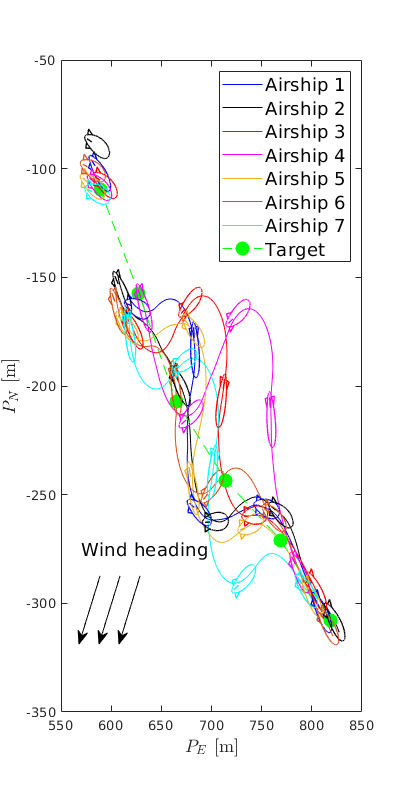}
        \caption{Detail B of the trajectory of random moving target and airship swarm using RPSO for 7 airships}
        \label{subfig:rpsodetailB7}
        \end{subfigure}
        \caption{Detailed views of the trajectory of 7 airships using RPSO, to follow a random moving target.}
        \label{fig:rpsoDetails7}       
\end{figure}
%
Table \ref{tab:errors} compiles the resulted average errors obtained across all the simulations, while table \ref{tab:std} presents the standard deviation of the same simulations.

\begin{table}[!htb]
\centering
\ra{1.1}
\caption{Average error across all simulations.}
\label{tab:errors}
\begin{tabular}{@{}lrr@{}}
\toprule
    & Waypoint Path [m] & Target Following [m] \\ \cmidrule(l){2-3}
    \bigcell{l}{Classical \\ Control} & 3.05 (followers) & \bigcell{r}{21.30 (leader), \\ 16.66 (followers)} \\  \addlinespace
    BOIDS & {2.95 (Swarm Centre)}  & {5.95 (Swarm Centre)} \\  \addlinespace
    RPSO  & {22.05 (Swarm Centre)} & {18.72 (Swarm Centre)} \\ 
 \bottomrule
\end{tabular}
\end{table}

\begin{table}[!htb]
\centering
\ra{1.1}
\caption{Standard error for table \ref{tab:errors}}
\label{tab:std}
\begin{tabular}{@{}lrr@{}}
\toprule
    & Waypoint Path [m] & Target Following [m] \\ \cmidrule(l){2-3}
    \bigcell{l}{Classical \\ Control} & 3.26 (followers) & \bigcell{r}{3.75 (leader), \\ 3.78 (followers)} \\  \addlinespace
    BOIDS & {1.03 (Swarm Centre)}  & {2.55 (Swarm Centre)} \\  \addlinespace
    RPSO  & {12.31 (Swarm Centre)} & {7.34 (Swarm Centre)} \\ 
 \bottomrule
\end{tabular}
\end{table}

\section{Discussion}
\label{sec:discussion}

Firstly, we should recall that the challenges related to both mission problems are many, as there should be simultaneous control of path tracking (either from a waypoint or a moving target) and the relative position between the agents, in the case of formation flight, or the entropy in the case of intelligent swarm. Further, the movement of the agents is a kind of "move-stop-move" walking that should be coordinated under strong wind/gust perturbations.

Also, some differences between the two control approaches should be remarked. First, in the case of formation flight, we have a kinematic guidance controller that sends reference commands to a nonlinear flight controller (SMC-based), both simulated using a dynamic airship model. However, for the intelligent swarm we have a direct kinematic controller applied to a kinematic airship model. Second, the intelligent swarm control has a natural "collision avoidance" scheme, while in the formation flight, in this first proposal, we considered the airships flying at different altitudes.

In the particular case of formation flight control, we noted the importance of a kinematic guidance layer, responsible for the leader/follower arrangement, in order to set distances and velocities references that are followed by the individual nonlinear controllers of the airships. 

Strong winds are always a challenge for airships, even in the single agent case, and moreover in a cooperative flight. Recommended guidelines for the design include increasing the distance between the agents, as well as the distance of the leader to the moving target in the ground tracking application. The scalability of the approach seems to be good, as we tested it for 4 and 7 agents. However, care should be taken, as the performance of a follower depends of the performance of its neighbour, which can yield cumulative degradation in tracking. The overall performance in the waypoint navigation is excellent, and similar to that of a single airship. In the moving target tracking case, the performance will depend on the escape velocity of the target. The worst combination is a sudden backward movement of the target in a tailwind condition (lower controllability), when temporary offset distance errors may appear.

For the Intelligent Swarm approach, both strategies, the Boids and the RPSO controllers, are able a efficiently drive a swarm of airships to perform the \textit{waypont} and to \textit{track a moving target} while remaining in formation. The Boids is the simplest one. It is based on a combination of artificial forces pulling each robot in a specific direction. Therefore the algorithm is easy to tune, e.g., to improve the repulsion effect, one just has to increase the constant parameter associated to it. However, this also means that these parameters have to be chosen carefully. For instance, if one is too high, the effect of its associated force may cancel the other. This is also true with the formulation of the forces. Sometimes it is better if one effect cancel the others, for example to avoid a collision.


Notwithstanding, the Boids controller is scalable to a large number airships. Even if the parameters (e.g., 9 constants in total) may have to be adjusted, the structure of the controller itself does not change, which is an advantage compared to classical formation controller which highly dependent on the number of robots.

The RPSO controller is more complex than the Boids, both to implement and to calibrate. The RPSO is based on an optimisation algorithm applied to robotics. The idea is that by moving the robots, the algorithm optimises a fitness function. We have chosen to optimise the entropy of the swarm and the distance to the target, so that the swarm would be more compact and would follow the target. This is both an advantageous and a disadvantageous compared to the Boids algorithm. For instance, the algorithm could be used to optimise as many variables as required, though the more parameters needed to be optimised, the longer it would take. However, this also means that the RPSO algorithm has no direct effect on the airships. The airships have to move to compare their fitness to their previous position’s one, and predict where to move at the next step. The algorithm has no information of what the optimal solution is. The Boids controller, on the other hand, has a direct effect on the robots, i.e., each of them is driven by the forces, which clearly indicate where to go the reach the target while avoiding the obstacles.

Therefore depending on the type of application, one could choose one or the other. Moreover, the RPSO algorithm may be more suitable for a foraging type behaviour. For instance, by removing the entropy from the values to optimise and adding the total area covered by the swarm, the RPSO algorithm could perform better than the Boids for an exploration or surveillance mission. The Boids controller induces a flocking behaviour, which is not compatible with a foraging behaviour and thus not extremely compatible with an exploration mission for which is more efficient to split the swarm to share the work.

\section{Conclusion}
\label{sec:conclusion}

In this paper we investigated the design and implementation of controllers for autonomous cooperative airships flights using two different approaches, namely formation control and swarm intelligence strategies. 

Two different swarm strategies have been tested to control the swarm of airships, one based on the Reynold’s Boids algorithm and the other on the Robotic Particle Swarm Optimisation algorithm. Both controllers can pilot the swarm to make it follow a pre-defined \textit{waypoint} and to \textit{track a moving target} while having a flocking behaviour and avoiding obstacles in a simulated 2D environment. The implementation and test of the different swarm strategies investigated have shown that both controllers were scalable to a large number of robots, as the exact same controller is implemented on each of them. It has also shown that the Boids inspired controller would perform better than the other on a mission where the goal can be directly formulated as a virtual force. 

Some disadvantages of the Boids and RPSO approaches have also been highlighted by this work. Indeed, both algorithms require the adjustment of a few parameters. Even if their effects are straightforward, it might take a quite long time to find a good set of parameters, which is usually valid only for a few specific conditions.


This work could be continued in many directions, as multiple areas linked to robotics have been approached. A strategy (or an algorithm) to adjust faster the parameters of the swarm strategy controllers would be a great improvement. For instance, this could be done with an optimisation algorithm which would launch a simulation, measure the performances of its parameters and adjust them accordingly.
Finally, the implementation of the simulator directly in ROS would allow the use of a more realistic simulation software, and thus the implementation of a more realistic obstacle detection system.



%


\appendix

\section*{Entropy computation algorithm}
\label{app:entropy}
\begin{algorithm}[H]
\small
	\caption{Entropy computation algorithm for a swarm of $N$ robots}
	\label{alg:entropy}
	\begin{algorithmic}
		\STATE \textbf{Inputs:} $3$-by-$N$ matrix $\mathbf{X}=\left[\mathbf{x}_1\cdots\mathbf{x}_N\right]$
		containing the positions of all the robots of the swarm
		\STATE \textbf{Output:} Entropy of the swarm $S$
		\STATE
		\STATE Initialise $S=0$
		\STATE Construct the $N$-by-$N$ matrix $\mathbf{D}$ containing the distances
		between each member of the swarm
		\begin{equation*}
		\mathbf{D}=\left[\begin{matrix}
		0 & D_{1,2} & \cdots & D_{1,N}\\
		D_{1,2} & 0 & & \vdots\\
		\vdots & & \ddots & D_{N-1,N}\\
		D_{1,N} & \cdots & D_{N-1,N} & 0
		\end{matrix}\right]
		\end{equation*}
		\begin{equation*}
		D_{i,j}=\|\mathbf{x}_i-\mathbf{x}_j\|
		\end{equation*}
		\STATE Suppress rows of $\mathbf{D}$, where agents distance is below a parameter $\delta_s$ which becomes a $N'$-by-$N$ matrix. Therefore, $N'$ is 
		the potential number of clusters of the swarm.
		\STATE Construct the 1-by-$n$ vector $\mathbf{d}$ containing all the different values of $\mathbf{D}$
		except $0$
		\FORALL{$h$ in $\mathbf{d}$}
			\STATE Create the $N'$-by-$N$ mask $\mathbf{M}$ of $\mathbf{D}$
			\begin{equation*}
			M_{i,j}=\begin{cases}
			1 & if\ D_{i,j}\leq h\\
			0 & \mathrm{otherwise}.
			\end{cases}
			\end{equation*}
			\STATE Compute $H(h)$
			\begin{align*}
			p_i &=\sum_{j=1}^{N} \frac{M_{i,j}}{N}, \quad \forall i \in [1,\ldots,N'],\\
			H(h) &=-\sum_{i=1}^{N'}\left(p_i\cdot\log_2\left(p_i\right)\right).
			\end{align*}
			\STATE Update $S$ by adding the value $h\cdot H(h)$
			\begin{equation*}
			S\leftarrow S+h\cdot H(h)
			\end{equation*}
		\ENDFOR
		\RETURN $S$
	\end{algorithmic}
\end{algorithm}

\emph{Remark.} As $p_i\in\left[\frac{1}{N},1\right]$, $\log_2\left(p_i\right)\leq 0$ and is always well-defined. Indeed, a cluster always contains at least one robot.


\section*{Acknowledgment}

The authors would like to thank FAPESP and Heriot-Watt University for the Sprint Funding (SAS-ROGE grant number 2016/50001-1). A. R. Fioravanti would also like to thank FAPESP, grant number 2018/04905-1, and CNPq, grant number 305600/2017-6. Ely Paiva acknowledges the fundings received from
FAPEAM (n. 253/2014, Dec. 014/2015), CNPq DRONI (n. 402112/2013-0), and Project INCT-SAC (CNPq 465755/2014-3, FAPESP 2014/50851-0).

\ifCLASSOPTIONcaptionsoff
  \newpage
\fi



\bibliographystyle{IEEEtran}
\bibliography{bibliography}
%



%

\begin{IEEEbiography}[{\includegraphics[width=1in,height=1.25in,clip,keepaspectratio]{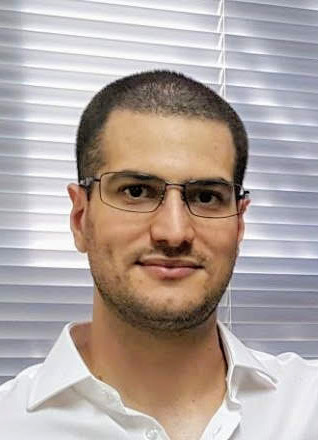}}]{Pedro G. Artaxo}
worked with robotic systems control, throughout his Undergraduate thesis in Control \& Automation Engineering (2016), to his Master's thesis in Mechanical Engineering (2018), both from the University of Campinas. Is currently enrolled in a Ph.D. on the same University, working with control in hybrid systems.
\end{IEEEbiography}

\begin{IEEEbiography}[{\includegraphics[width=1in,height=1.25in,clip,keepaspectratio]{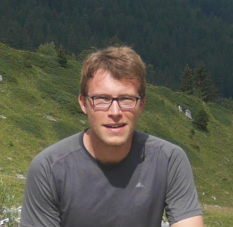}}]{Auguste Bourgois}
 is currently working as a research engineer at Forssea Robotics in Paris, France, and is writing his PhD. thesis on autonomous underwater docking at ENSTA Bretagne, Brest, France. He obtained his MSc. in Robotics and Autonomous Systems from Heriot-Watt univertsity, Edinburgh, UK, and his French engineering diploma in Mobile Robotics from ENSTA Bretagne in 2017. His work interests lie in the field of marine robotics, ranging from robot design and software architecture to robot localisation, control and navigation.
\end{IEEEbiography}


\begin{IEEEbiography}[{\includegraphics[width=1in,height=1.25in,clip,keepaspectratio]{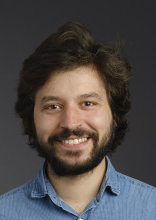}}]{Hugo Sardinha}
received his MSc in Mechanical Engineering with a specialism in Control Systems, from Instituto Superior T\'ecnico (IST), Lisbon University in 2012. After working for 3 years industry he rejoined academia and received a MSc in Robotics, Autonomous and Interactive Systems from Heriot-Watt University, in 2016. Currently he is a PhD candidate at the Edinburgh Centre for Robotics working on adaptive behaviours for aerial swarm robotics.
\end{IEEEbiography}

\begin{IEEEbiography}[{\includegraphics[width=1in,height=1.25in,clip,keepaspectratio]{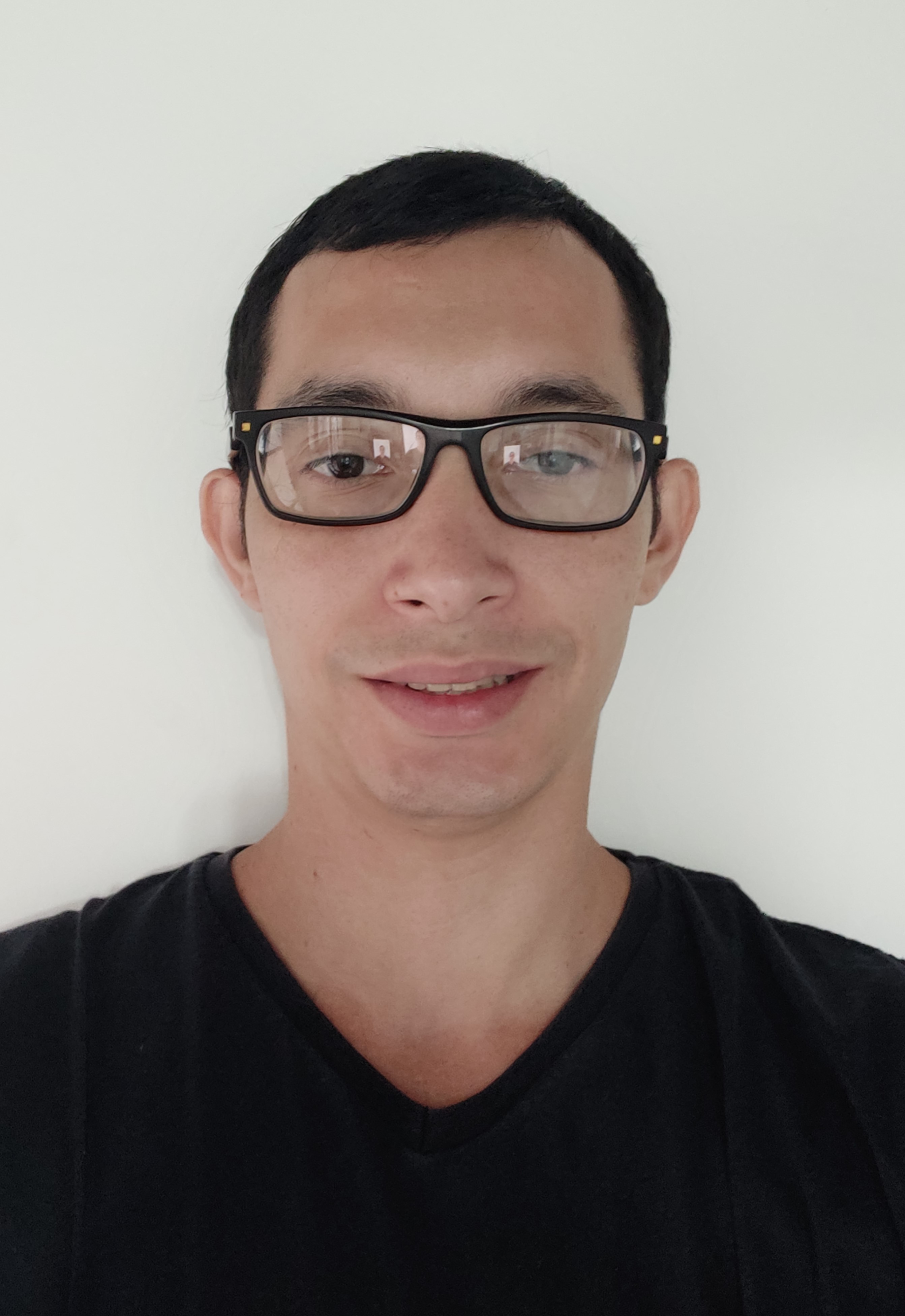}}]{Henrique Vieira}
was born in Brazil, in 1989. He has a Bachelor degree in Mechatronics Engineering from Amazonas State University (2012), Masters in Electrical Engineering from Campinas State University (2015). Ph.D. in Mechanical Engineering in Campinas State University (2019), in the field of nonlinear control design with applications to an autonomous airship.
\end{IEEEbiography}

\begin{IEEEbiography}
[{\includegraphics[width=1in,height=1.25in,clip,keepaspectratio]{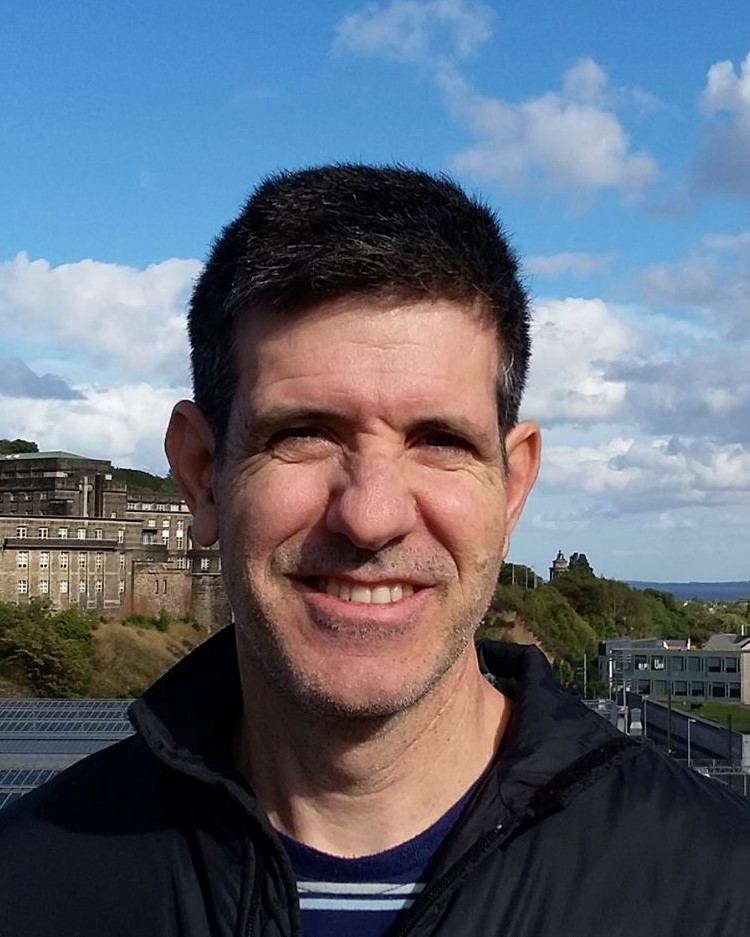}}]{Ely C. de Paiva}
was born in Brazil, in 1965. He
received the Ph.D. degree in Electrical Engineering
(Automation) in the University of Campinas (Unicamp),
Brazil, in 1997. From 1997 until 2009, he
worked as researcher of the Center
for Information Technology Renato Archer, Campinas, Brazil,
within the Robotics and Computer Vision Division.
Since 2010, he is professor of the School of
Mechanical Engineering of Unicamp. In 2018/2019 he was in  research leave at Concordia University of Montreal, working with Optimal Guidance and Motion Planning for unmanned aerial vehicles (UAVs). His research interests include robust and nonlinear
control, mobile robotics, autonomous vehicles,  modelling and flight control,
\end{IEEEbiography}

\begin{IEEEbiography}[{\includegraphics[width=1in,height=1.25in,clip,keepaspectratio]{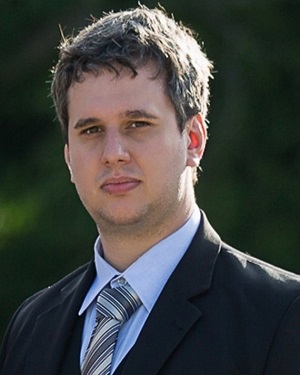}}]{Andr\'{e} R. Fioravanti}
was born in Brazil, in 1982. He received his B.Sc and M.Sc degrees in Electrical Engineering from University of Campinas, Brazil (2007, 2008), and the Ph.D. degree in Physics from Universit\'{e} Paris-Sud XI, France (2011). He is currently an Assistant Professor at Univeristy of Campinas, Brazil and class 2 researcher from the Brazilian National Research Center (CNPq). His research spans from control and dynamical systems to optimization methods and simulation. He focuses on questions about stability and performance of dynamical systems, decision making through mathematical programming and complex systems simulation.
\end{IEEEbiography}

\begin{IEEEbiography}[{\includegraphics[width=1in,height=1.25in,clip,keepaspectratio]{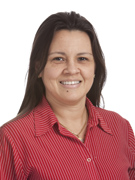}}]{Patricia A. Vargas}

is the Founder Director of the Robotics Laboratory within the Edinburgh Centre for Robotics and Associate Professor/Reader in Robotics and Computer Science at Heriot–Watt University, Edinburgh, UK. She is a member of the Higher Education Academy, IEEE Robotics and Automation, and Computational Intelligence Societies. Patricia received her PhD on Computer Engineering from the University of Campinas, Unicamp (Brazil) in 2005. She was a post-doc at the Centre for Computational Neuroscience and Robotics, University of Sussex in England, UK, for 3 years. Her research interests include but are not restricted to Evolutionary and Bio-inspired Robotics, Swarm Robotics, Computational Neuroscience, Deep Learning, Human–Robot Interaction, Rehabilitation Robotics and Neurorobotics.
\end{IEEEbiography}

\vfill



\end{document}